\documentclass[10pt,twocolumn,letterpaper]{article}

\usepackage{cvpr}              

\usepackage{graphicx}
\usepackage{amsmath}
\usepackage{amssymb}
\usepackage{booktabs}
\usepackage{dcolumn,booktabs}
\newcolumntype{d}[1]{D{.}{.}{#1}}
\newcommand\mc[1]{\multicolumn{1}{c}{#1}} 

\usepackage[pagebackref,breaklinks,colorlinks]{hyperref}

\usepackage[capitalize]{cleveref}
\crefname{section}{Sec.}{Secs.}
\Crefname{section}{Section}{Sections}
\Crefname{table}{Table}{Tables}
\crefname{table}{Tab.}{Tabs.}

\def\cvprPaperID{2} 
\def\confName{CVPR}
\def\confYear{2024}

\hypersetup{colorlinks,linkcolor={blue},citecolor={blue},urlcolor={red}}

\begin{document}

\title{Finding AI-Generated Faces in the Wild}

\author{Gonzalo J. Aniano Porcile$^{1}$, Jack Gindi$^{1}$, Shivansh Mundra$^{1}$, James R. Verbus$^{1}$, Hany Farid$^{1,2}$\\
LinkedIn$^{1}$ and University of California, Berkeley$^{2}$\\
Sunnyvale CA, USA$^{1}$ and Berkeley CA, USA$^{2}$\\
{\tt\small \{ganiano,jgindi,smundra,jverbus\}@linkedin.com} and {\tt\small hfarid@berkeley.edu}
}
\maketitle


\begin{abstract}
AI-based image generation has continued to rapidly improve, producing increasingly more realistic images with fewer obvious visual flaws. AI-generated images are being used to create fake online profiles which in turn are being used for spam, fraud, and disinformation campaigns. As the general problem of detecting any type of manipulated or synthesized content is receiving increasing attention, here we focus on a more narrow task of distinguishing a real face from an AI-generated face. This is particularly applicable when tackling inauthentic online accounts with a fake user profile photo. We show that by focusing on only faces, a more resilient and general-purpose artifact can be detected that allows for the detection of AI-generated faces from a variety of GAN- and diffusion-based synthesis engines, and across image resolutions (as low as $128 \times 128$ pixels) and qualities.
\end{abstract}

\section{Introduction}
\label{sec:intro}

\begin{figure}[t]
    \begin{center}
        \begin{tabular}{c@{\hspace{0.15cm}}c}
          (a) & (b) \\
          \includegraphics[width=0.22\textwidth]{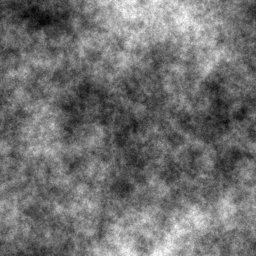} &
          \includegraphics[width=0.22\textwidth]{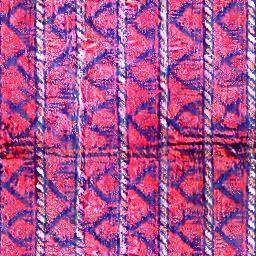} \\
          (c) & (d) \\
          \includegraphics[width=0.22\textwidth]{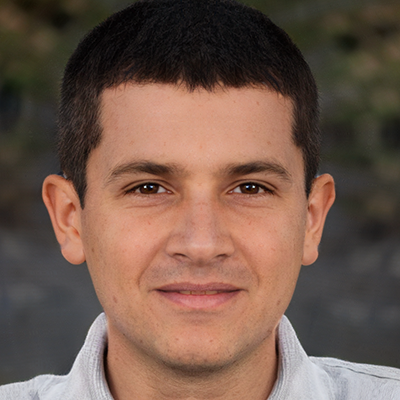} &
          \includegraphics[width=0.22\textwidth]{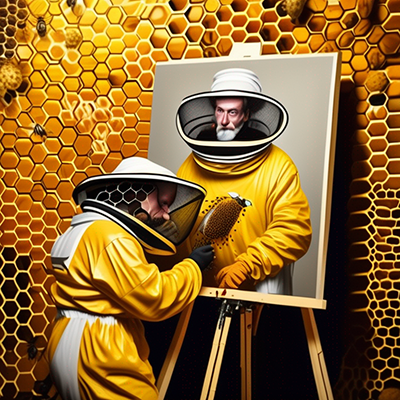}
        \end{tabular}
    \end{center}
    \vspace{-0.5cm}
    \caption{The evolution of statistical models of natural images: (a) a fractal pattern with a $1/\omega$ power spectrum; (b) a synthesized textile pattern\cite{portilla2000parametric}; (c) a GAN-generated face~\cite{karras2021alias}; and (d) a diffusion-generated scene with the prompt ``a beekeeper painting a self portrait''~\cite{stabilityAI}.}
    \label{fig:image-models}
\end{figure}

The past three decades have seen remarkable advances in the statistical modeling of natural images. The simplest power-spectral model~\cite{knill1990human} captures the $1/\omega$ frequency magnitude fall-off typical of natural images, Figure~\ref{fig:image-models}(a). Because this model does not incorporate any phase information, it is unable to capture detailed structural information. By early 2000, new statistical models were able to capture the natural statistics of both magnitude and (some) phase~\cite{portilla2000parametric}, leading to breakthroughs in modeling basic texture patterns, Figure~\ref{fig:image-models}(b). 

While able to capture repeating patterns, these models are not able to capture the geometric properties of objects, faces, or complex scenes. Starting in 2017, and powered by large data sets of natural images, advances in deep learning, and powerful GPU clusters, generative models began to capture detailed properties of human faces and objects
~\cite{karras2017progressive,karras2019style}. Trained on a large number of images from a single category (faces, cars, cats, etc.), these generative adversarial networks (GANs) capture highly detailed properties of, for example, faces, Figure~\ref{fig:image-models}(c), but are constrained to only a single category. Most recently, diffusion-based models~\cite{bau2021paint,rombach2022high} have combined generative image models with linguistic prompts allowing for the synthesis of images from descriptive text prompts like ``a beekeeper painting a self portrait'', Figure~\ref{fig:image-models}(d).

Traditionally, the development of generative image models were driven by two primary goals: (1) understand the fundamental statistical properties of natural images; and (2) use the resulting synthesized images for everything from computer graphics rendering to human psychophysics and data augmentation in classic computer vision tasks. Today, however, generative AI has found more nefarious use cases ranging from spam to fraud and additional fuel for disinformation campaigns. 

Detecting manipulated or synthesized images is particularly challenging when working on large-scale networks with hundreds of millions of users. This challenge is made even more significant when the average user struggles to distinguish a real from a fake face~\cite{nightingale2022ai}. Because we are concerned with the use of generative AI in creating fake online user accounts, we seek to develop fast and reliable techniques that can distinguish real from AI-generated faces. We next place our work in context of related techniques.

\subsection{Related Work}

Because we will focus specifically on AI-generated faces, we will review related work also focused on, or applicable to, distinguishing real from fake faces. There are two broad categories of approaches to detecting AI-generated content~\cite{farid2022creating}.

In the first, hypothesis-driven approaches, specific artifacts in AI-generated faces are exploited such as  inconsistencies in bilateral facial symmetry in the form of corneal reflections~\cite{guo2022eyes} and pupil shape~\cite{hu2021exposing}, or inconsistencies in head pose and the spatial layout of facial features (eyes, tip of nose, corners of mouth, chin, etc.)~\cite{yang2019exposingheadpose,yang2019exposinglandmarks,mundra2023exposing}. The benefit of these approaches is that they learn explicit, semantic-level anomalies. The drawback is that over time synthesis engines appear to be -- either implicitly or explicitly -- correcting for these artifacts. Other non-face specific artifacts include spatial frequency or noise anomalies~\cite{zhang2019detecting,chai2020makes,gragnaniello2021gan,liu2022detecting,corvi2023detection}, but these artifacts tend to be vulnerable to simple laundering attacks (e.g.,~transcoding, additive noise, image resizing).

In the second, data-driven approaches, machine learning is used to learn how to distinguish between real and AI-generated images~\cite{wang2020cnn,frank2020leveraging,tan2023learning,ju2023glff}. These models often perform well when analyzing images consistent with their training, but then struggle with out-of-domain images and/or are vulnerable to laundering attacks because the model latches onto low-level artifacts~\cite{dong2022think}.

We attempt to leverage the best of both of these approaches. By training our model on a range of synthesis engines (GAN and diffusion), we seek to avoid latching onto a specific low-level artifact that do not generalize or may be vulnerable to simple laundering attacks. By focusing on only detecting AI-generated faces (and not arbitrary synthetic images), we show that our model seems to have captured a semantic-level artifact distinct to AI-generated faces which is highly desirable for our specific application of finding potentially fraudulent user accounts. We also show that our model is resilient to detecting AI-generated faces not previously seen in training, and is resilient across a large range of image resolutions and qualities.

\section{Data sets}
\label{sec:datasets}

\begin{figure*}[t]
    \begin{center}
    \begin{tabular}{rc@{\hspace{0.15cm}}c@{\hspace{0.15cm}}c@{\hspace{0.15cm}}c@{\hspace{0.15cm}}c@{\hspace{0.15cm}}c}
         \raisebox{0.9cm}{generated.photos} &
         \includegraphics[width=0.11\textwidth]{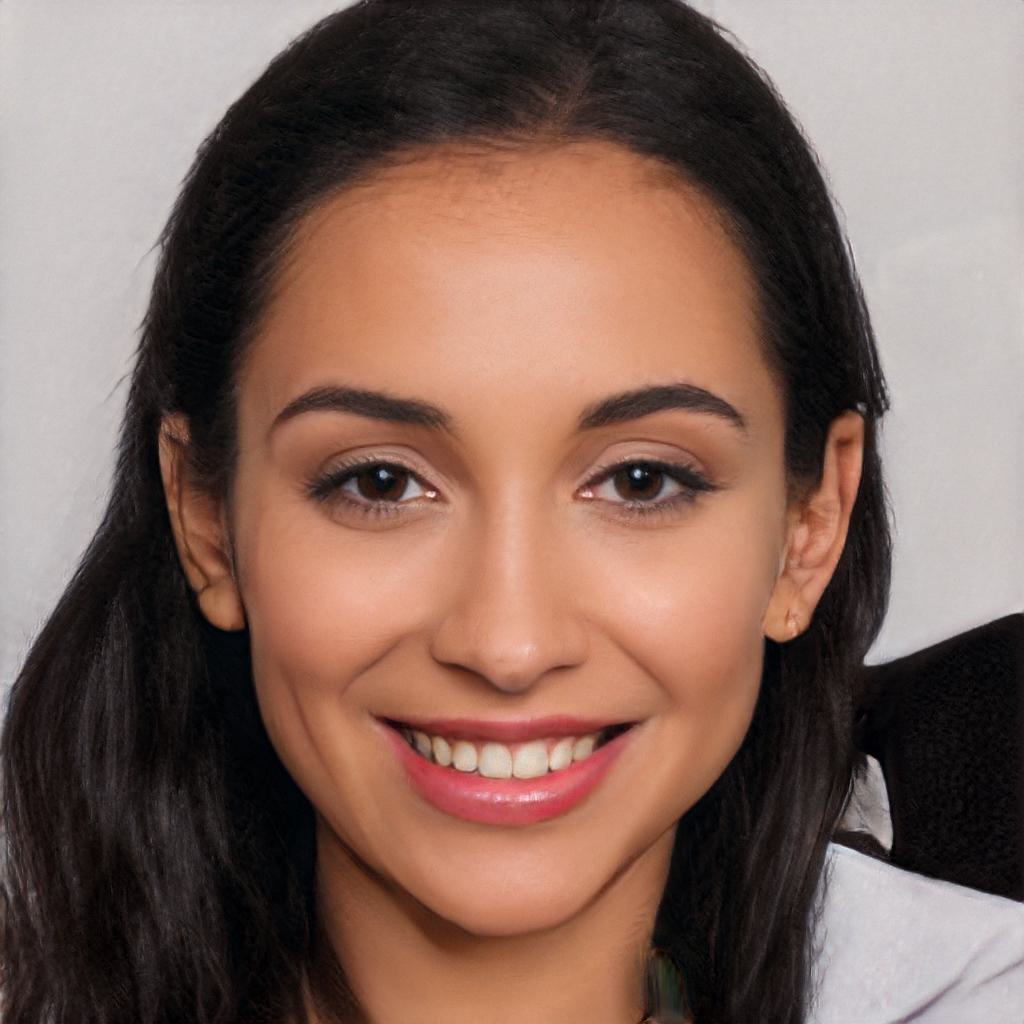} &
         \includegraphics[width=0.11\textwidth]{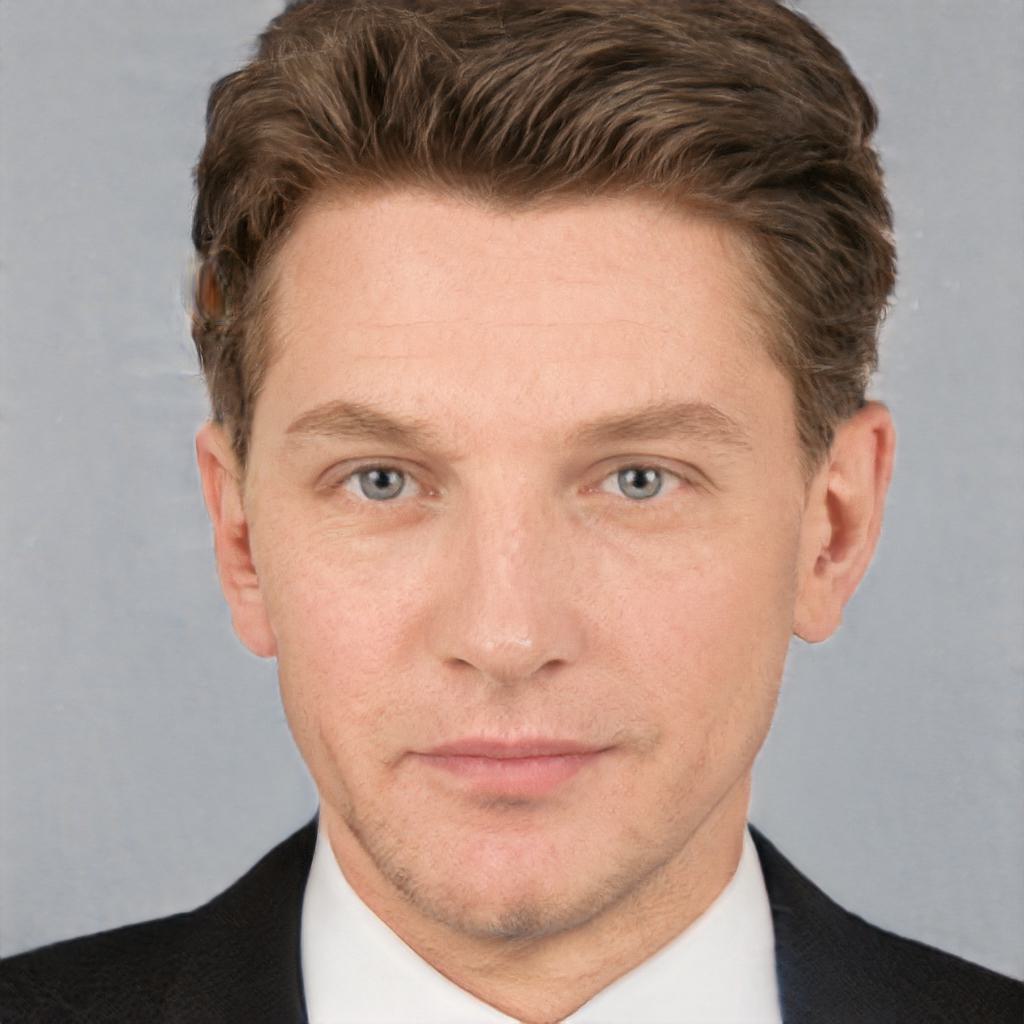} &
         \includegraphics[width=0.11\textwidth]{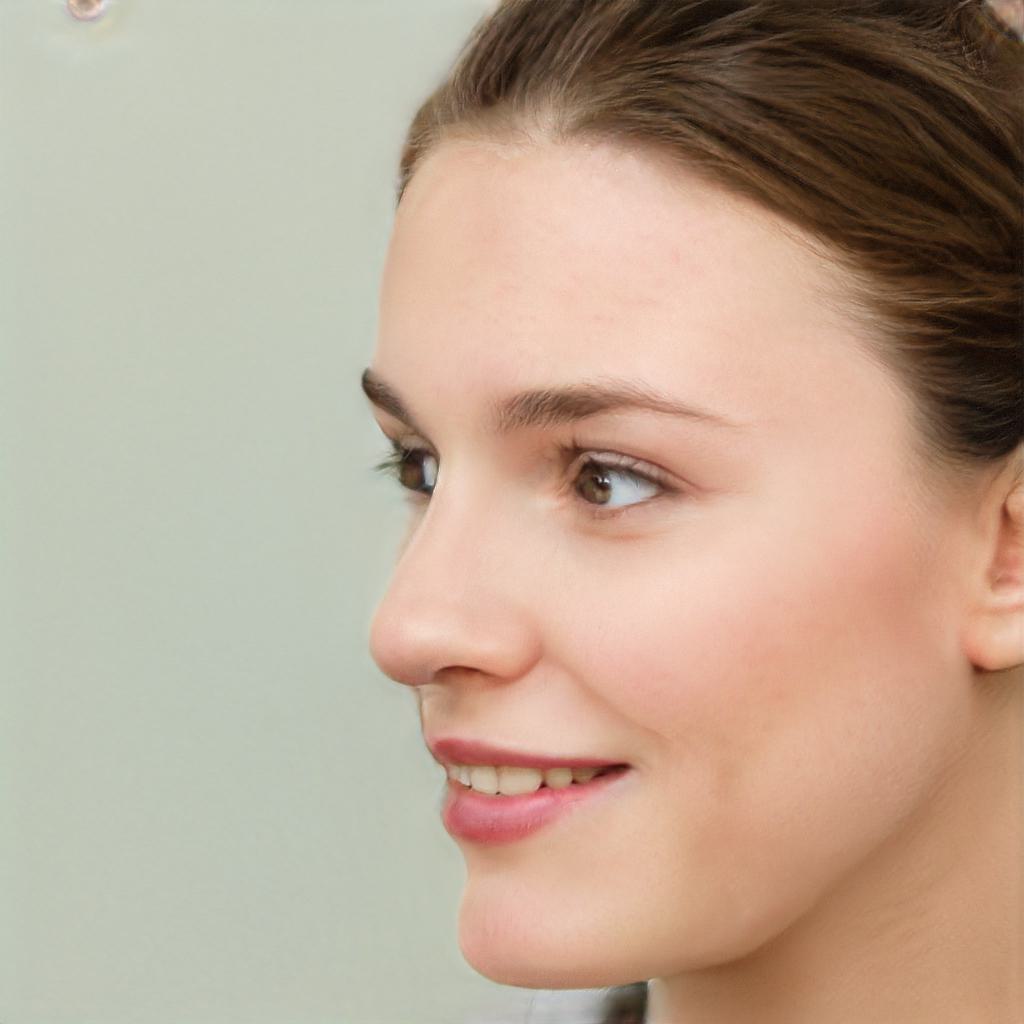} &
         \includegraphics[width=0.11\textwidth]{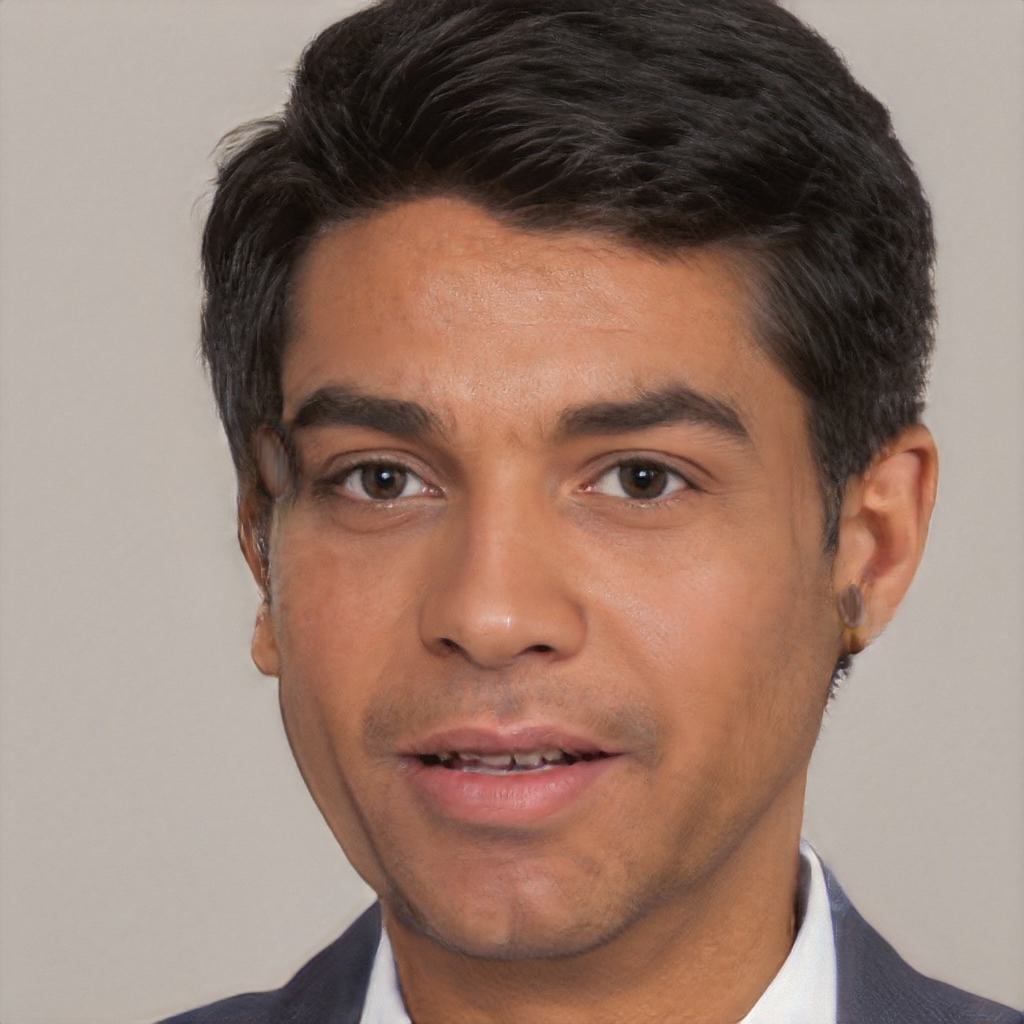} &
         \includegraphics[width=0.11\textwidth]{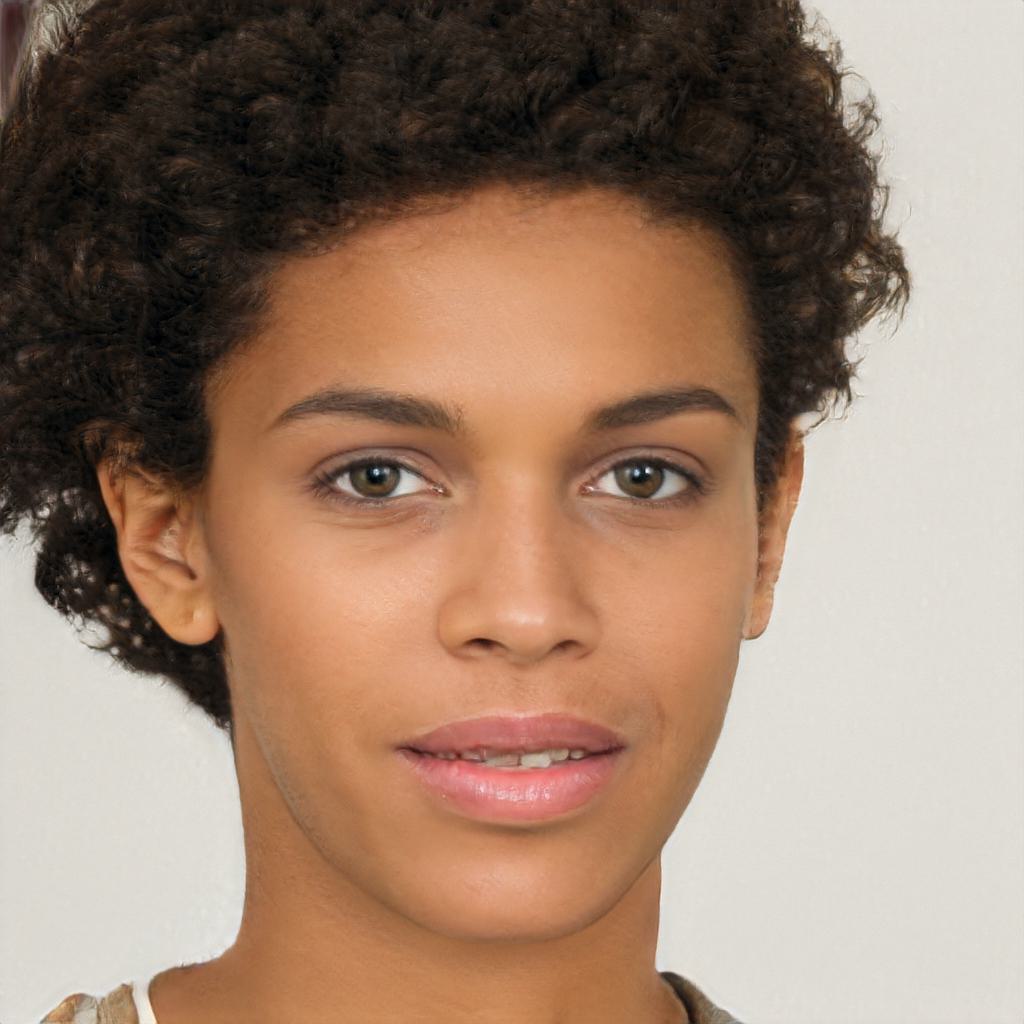} &
         \includegraphics[width=0.11\textwidth]{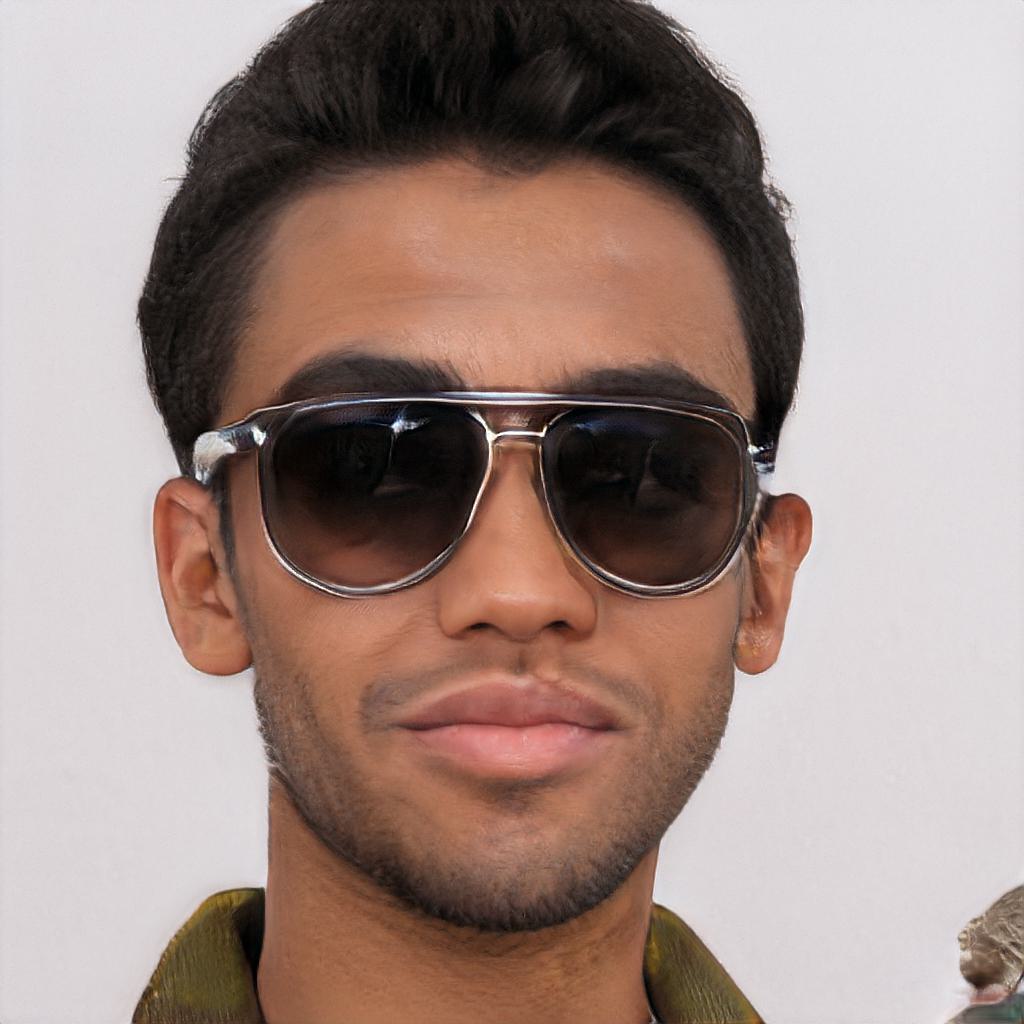}
        \\
        \raisebox{0.9cm}{StyleGAN 1} &
        \includegraphics[width=0.11\textwidth]{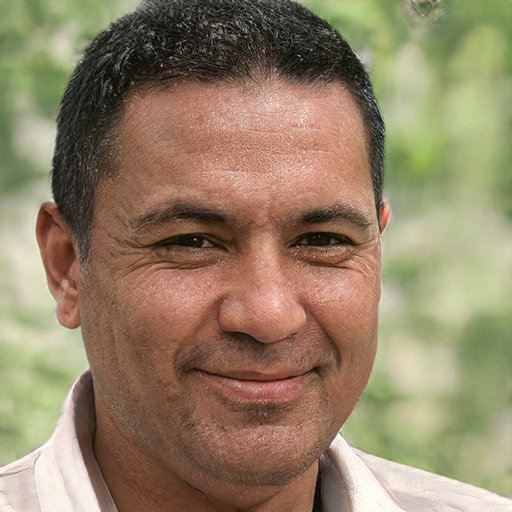} &
        \includegraphics[width=0.11\textwidth]{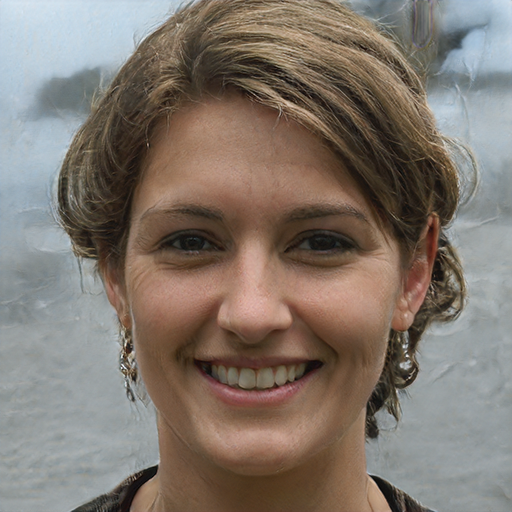} &
        \includegraphics[width=0.11\textwidth]{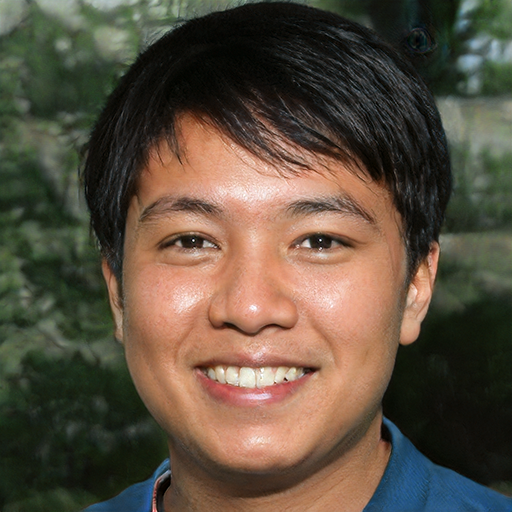} &
        \includegraphics[width=0.11\textwidth]{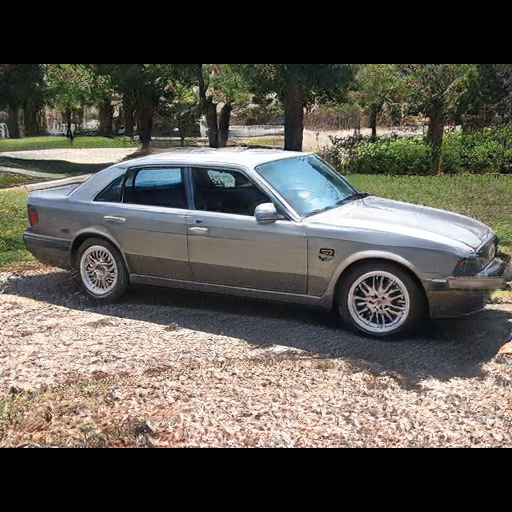} &
        \includegraphics[width=0.11\textwidth]{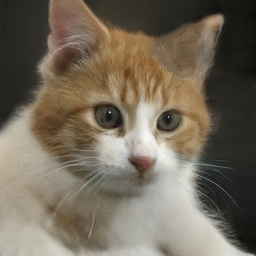} & 
        \includegraphics[width=0.11\textwidth]{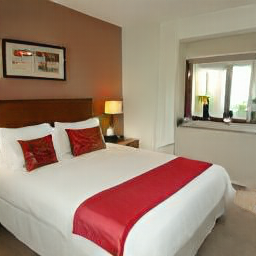} 
        \\
        \raisebox{0.9cm}{StyleGAN 2} & 
        \includegraphics[width=0.11\textwidth]{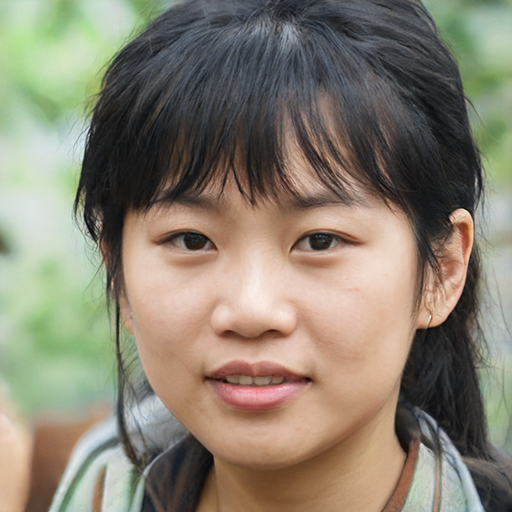} &
        \includegraphics[width=0.11\textwidth]{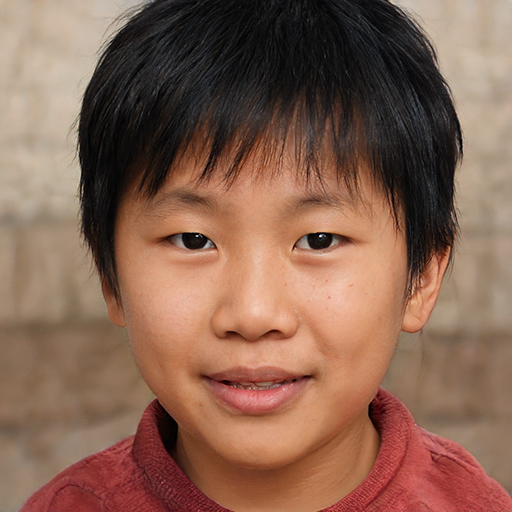} &
        \includegraphics[width=0.11\textwidth]{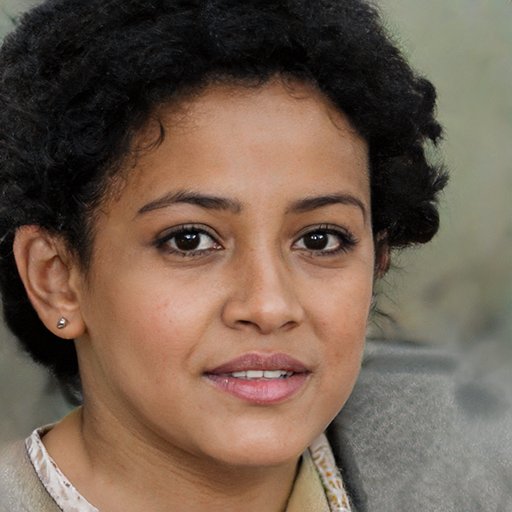} &
        \includegraphics[width=0.11\textwidth]{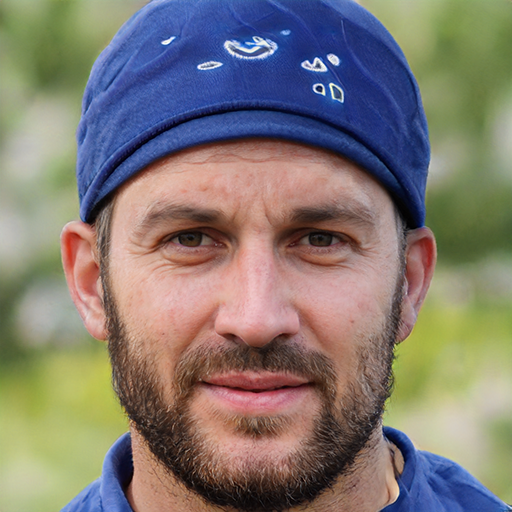} &
        \includegraphics[width=0.11\textwidth]{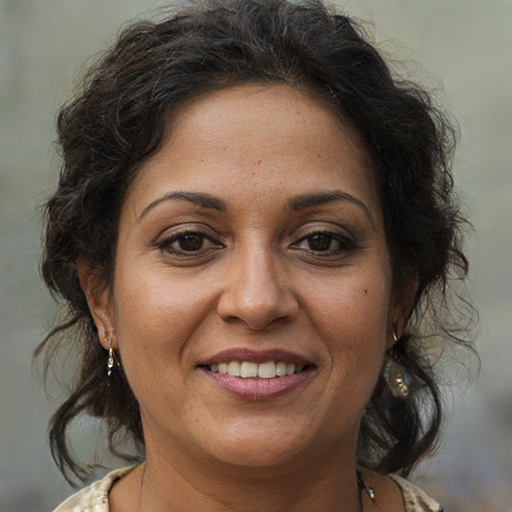} &
        \includegraphics[width=0.11\textwidth]{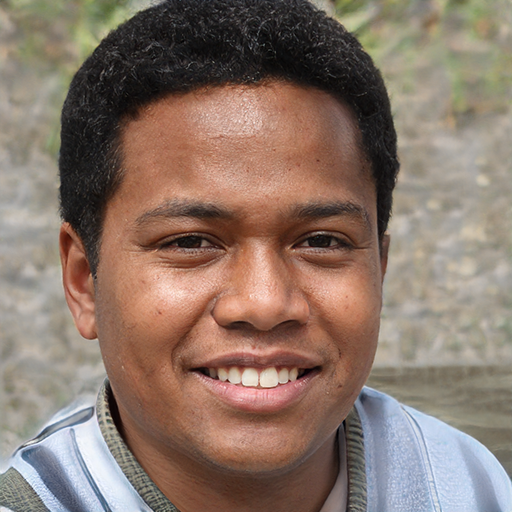}
        \\
        \raisebox{0.9cm}{StyleGAN 3} &
        \includegraphics[width=0.11\textwidth]{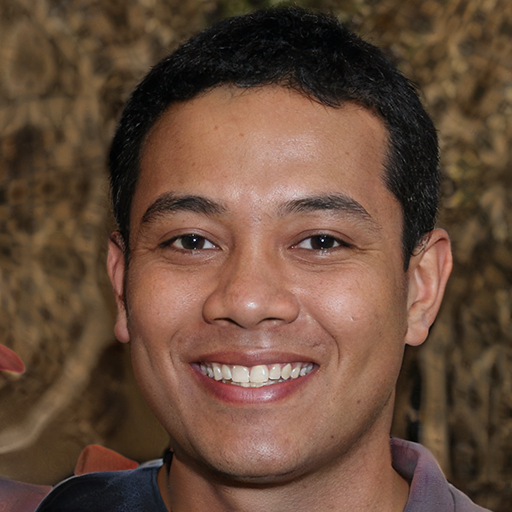} &
        \includegraphics[width=0.11\textwidth]{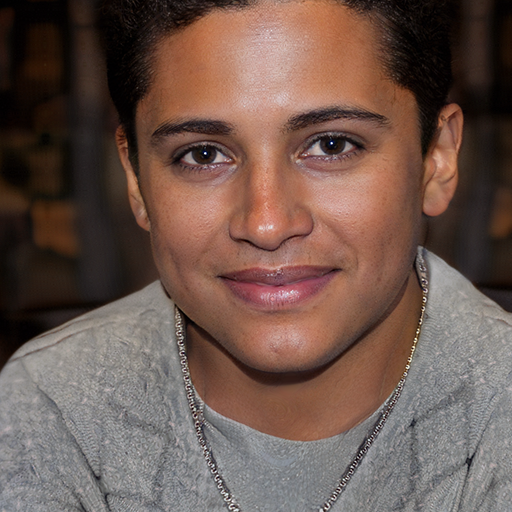} &
        \includegraphics[width=0.11\textwidth]{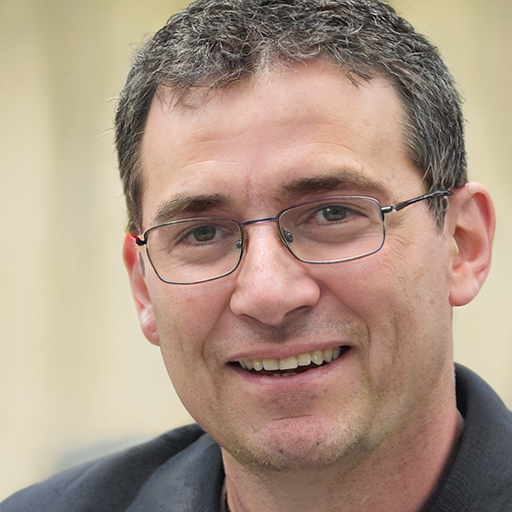} &
        \includegraphics[width=0.11\textwidth]{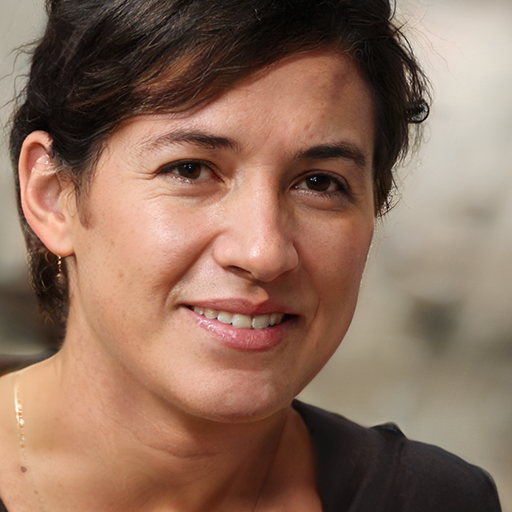} &
        \includegraphics[width=0.11\textwidth]{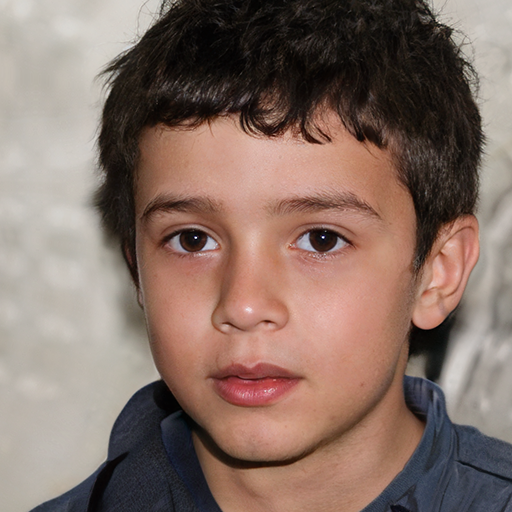} &
        \includegraphics[width=0.11\textwidth]{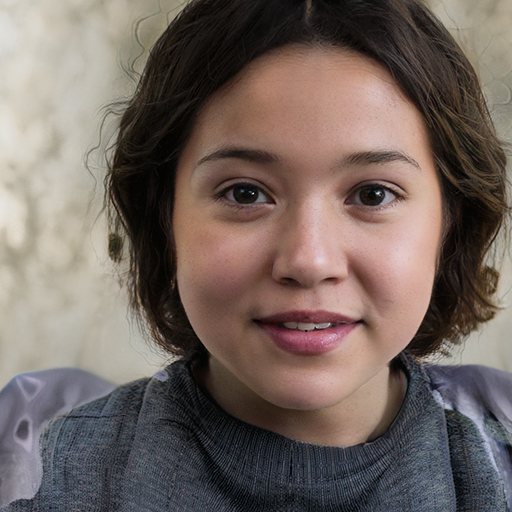} 
        \\
        \raisebox{0.9cm}{EG3D} &
        \includegraphics[width=0.11\textwidth]{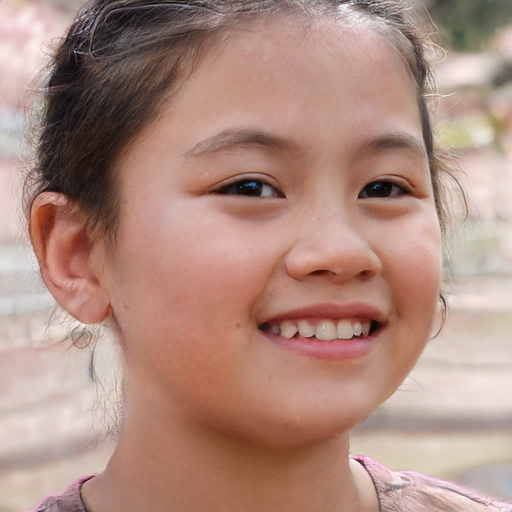} &
        \includegraphics[width=0.11\textwidth]{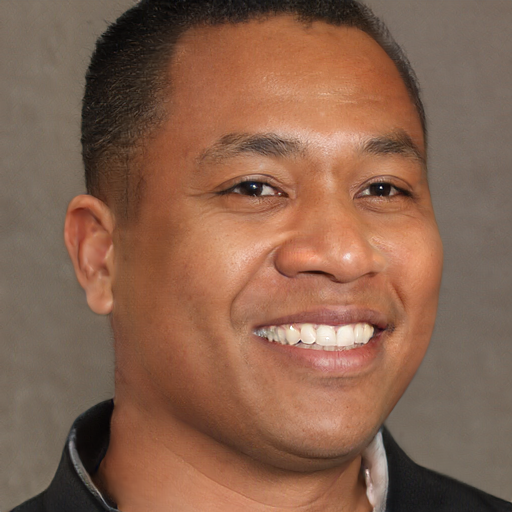} &
        \includegraphics[width=0.11\textwidth]{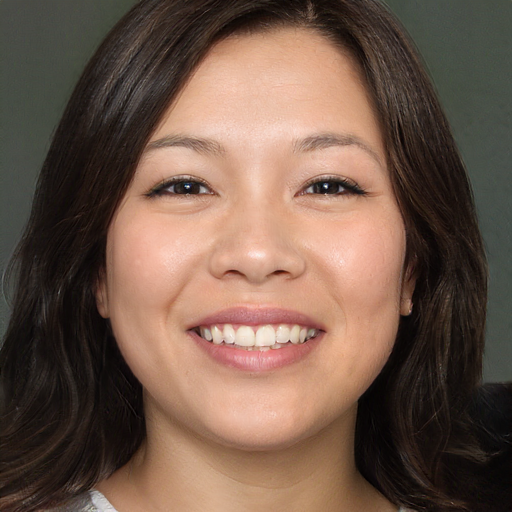} &
        \includegraphics[width=0.11\textwidth]{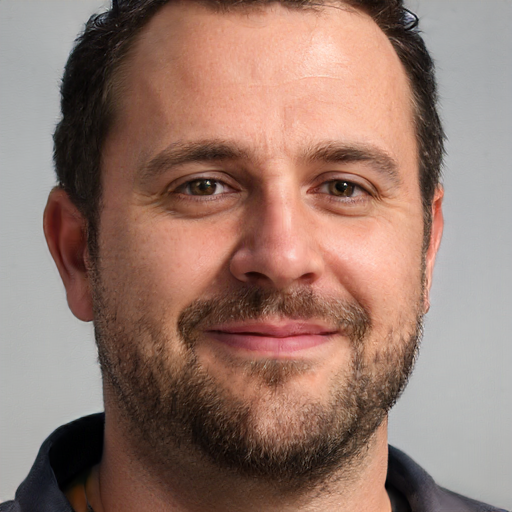} &
        \includegraphics[width=0.11\textwidth]{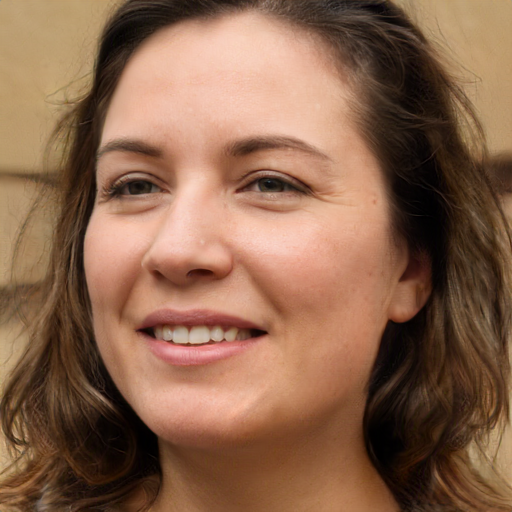} &
        \includegraphics[width=0.11\textwidth]{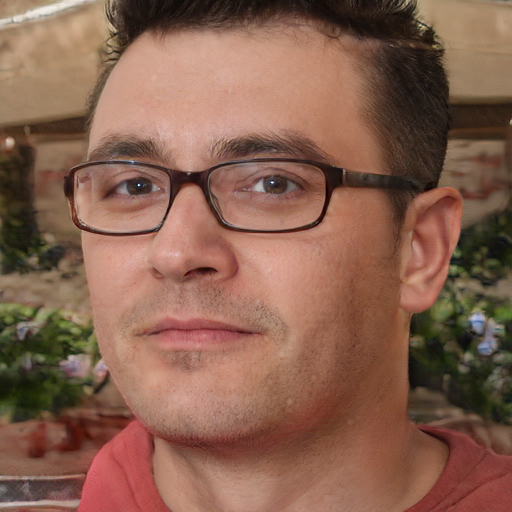}
        \\
        \raisebox{0.9cm}{DALL-E 2} & 
        \includegraphics[width=0.11\textwidth]{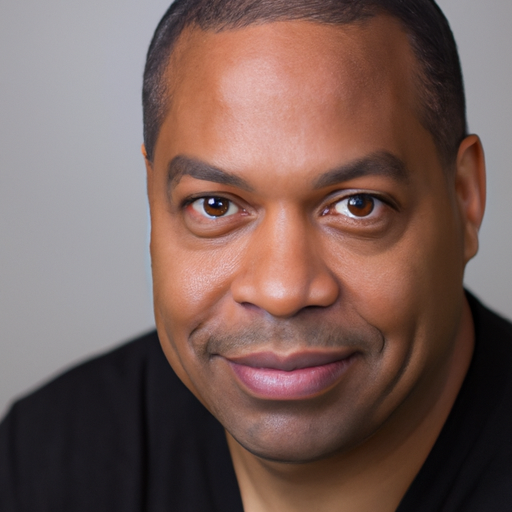} &
        \includegraphics[width=0.11\textwidth]{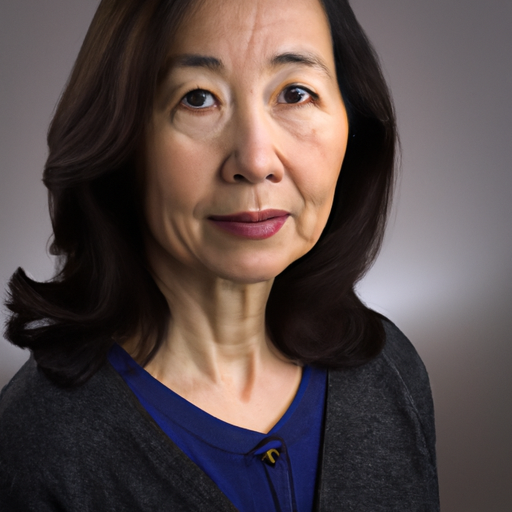} &
        \includegraphics[width=0.11\textwidth]{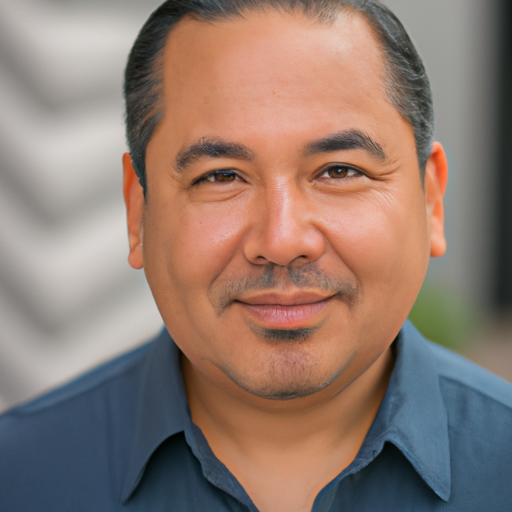} &
        \includegraphics[width=0.11\textwidth]{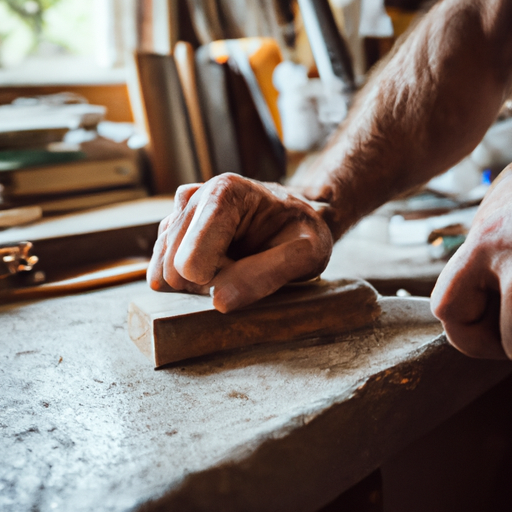} &
        \includegraphics[width=0.11\textwidth]{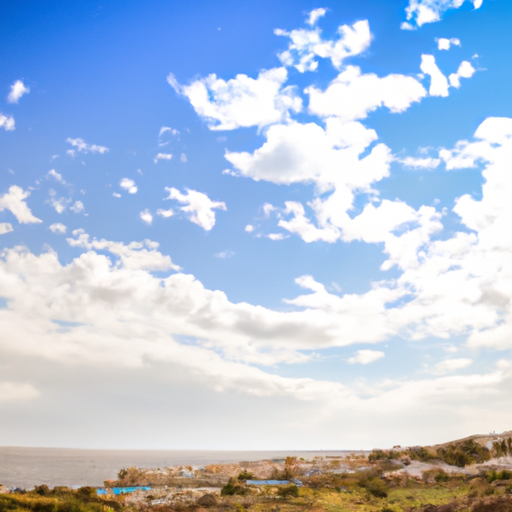} &
        \includegraphics[width=0.11\textwidth]{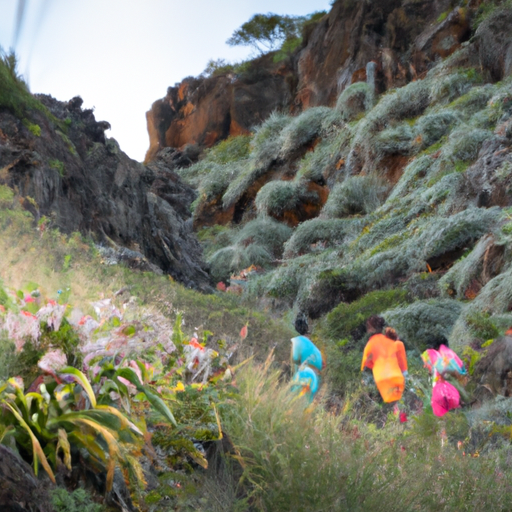}
        \\
        \raisebox{0.9cm}{Midjourney} & 
        \includegraphics[width=0.11\textwidth]{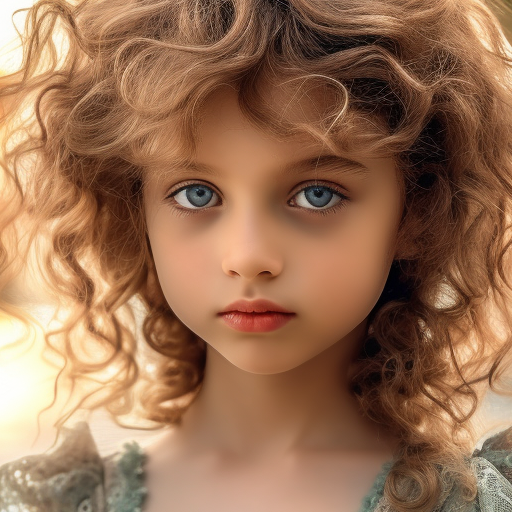} &
        \includegraphics[width=0.11\textwidth]{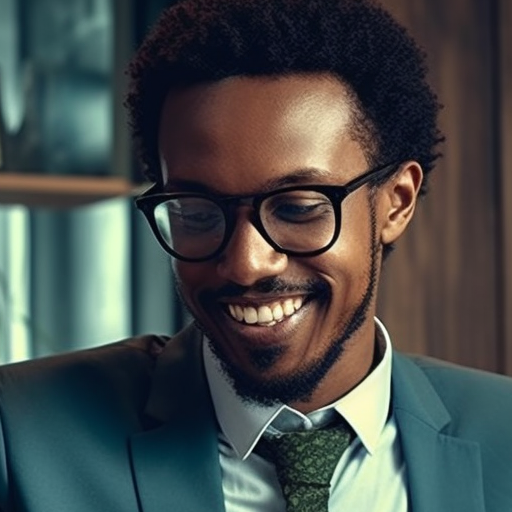} &
        \includegraphics[width=0.11\textwidth]{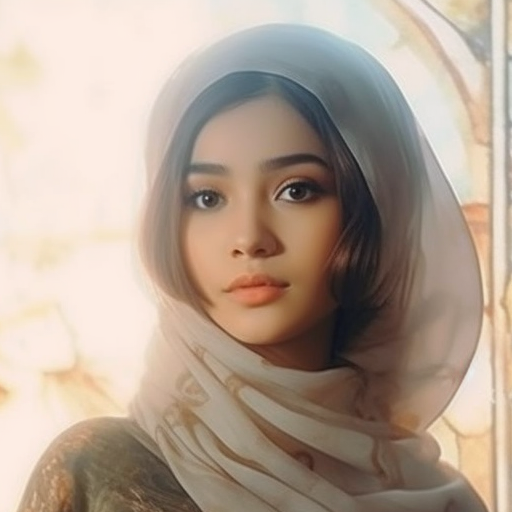} &
        \includegraphics[width=0.11\textwidth]{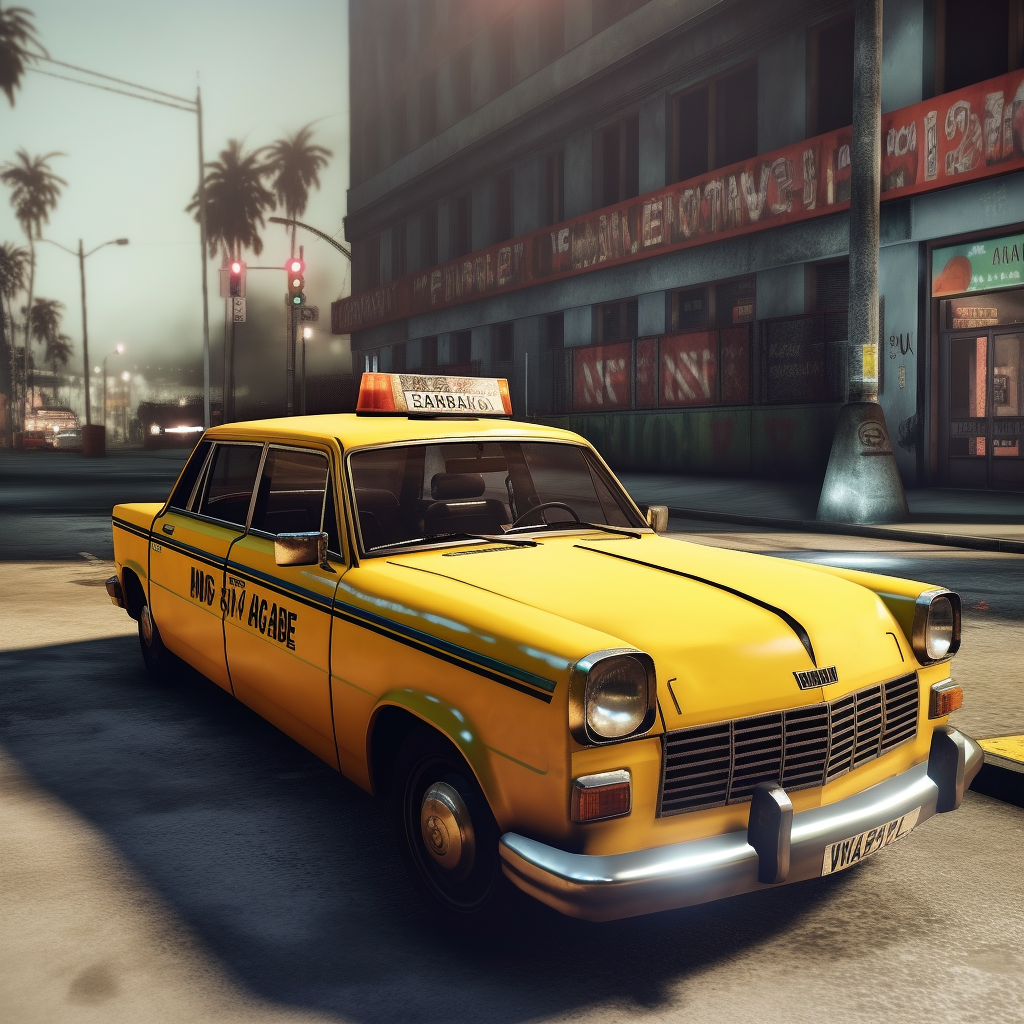} &
        \includegraphics[width=0.11\textwidth]{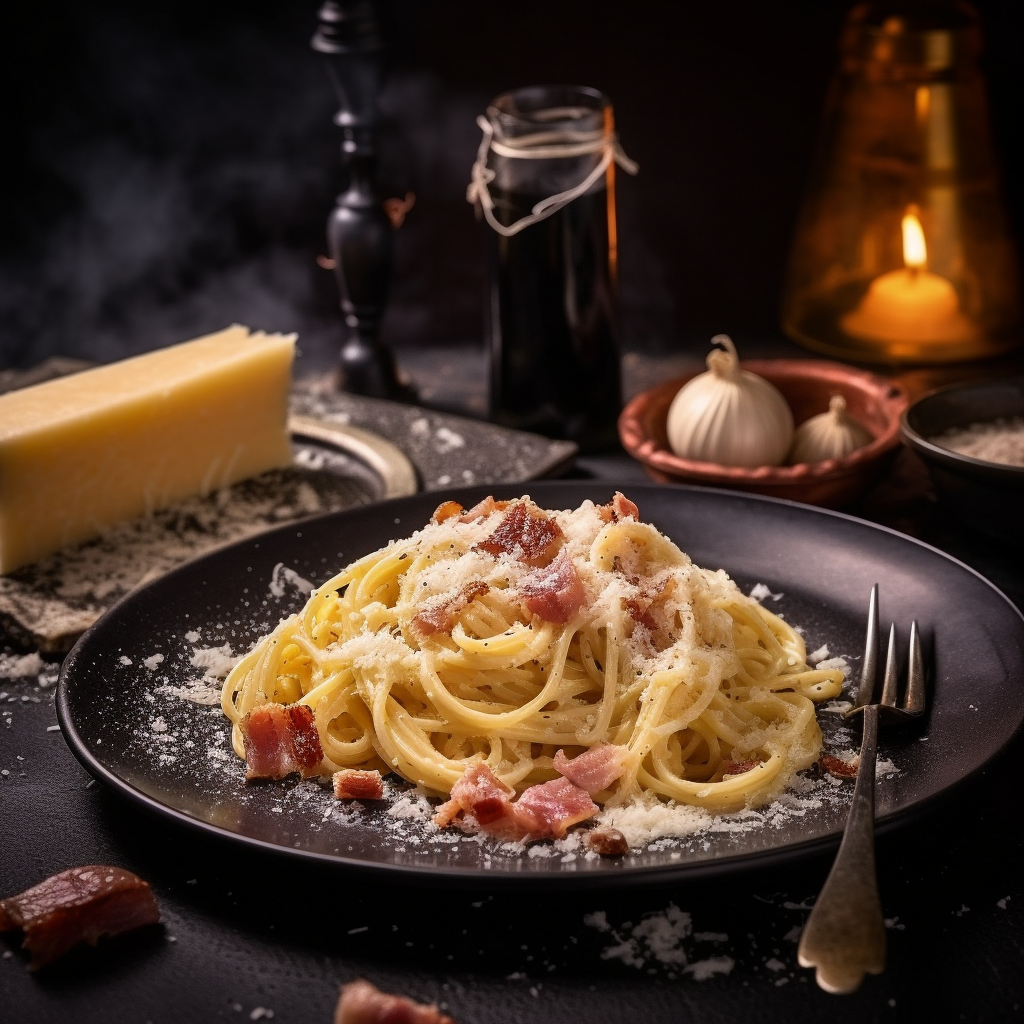} &
        \includegraphics[width=0.11\textwidth]{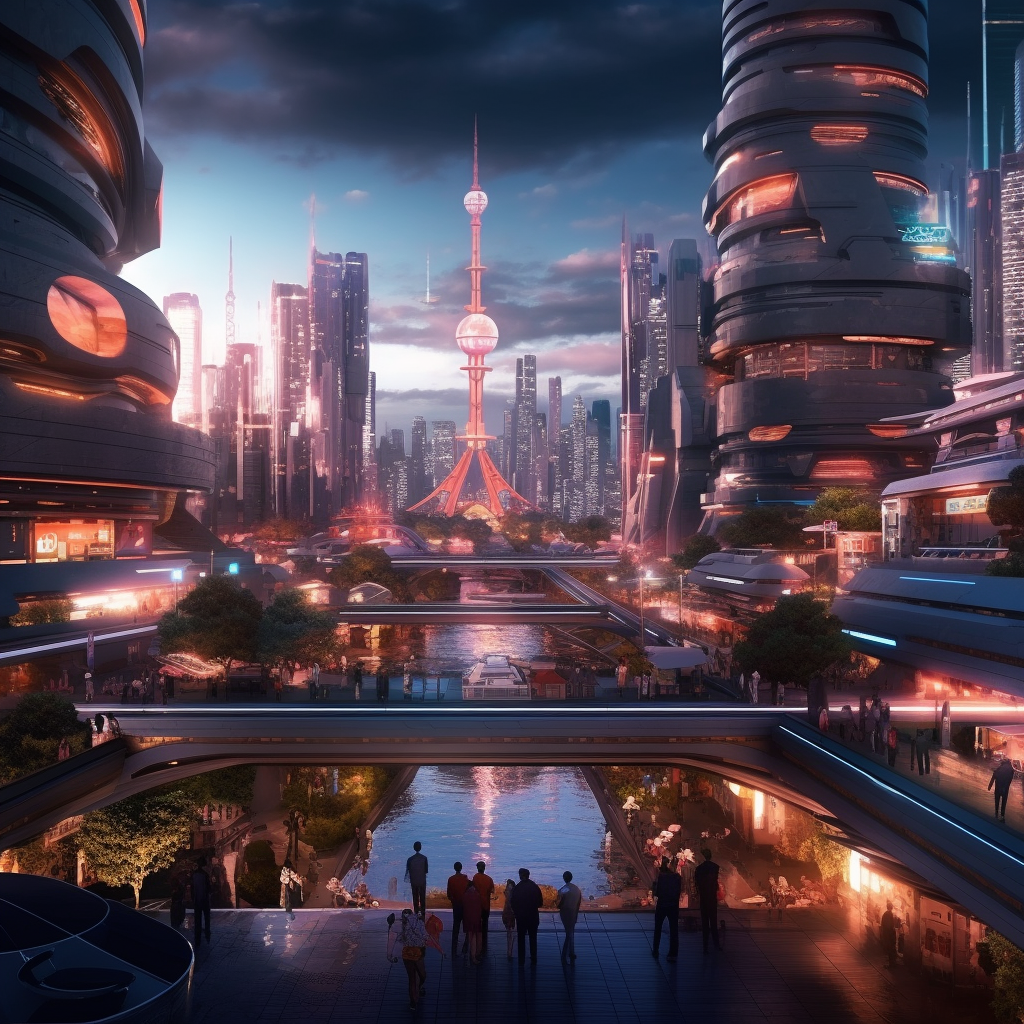}
        \\
        \raisebox{0.9cm}{Stable Diffusion 1} & 
        \includegraphics[width=0.11\textwidth]{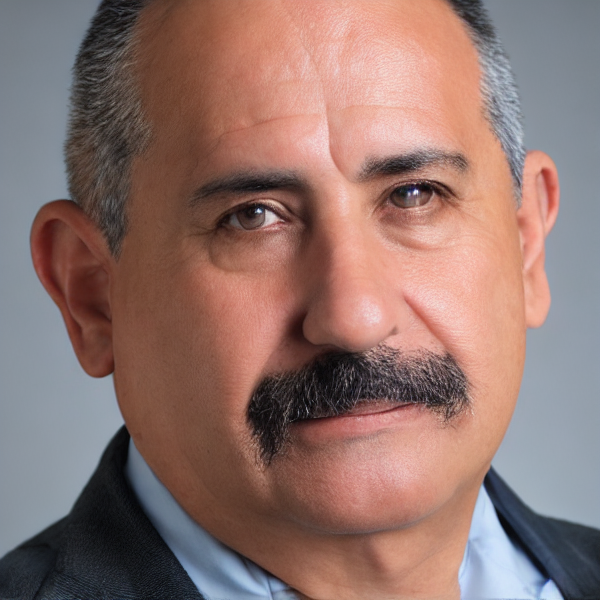} &
        \includegraphics[width=0.11\textwidth]{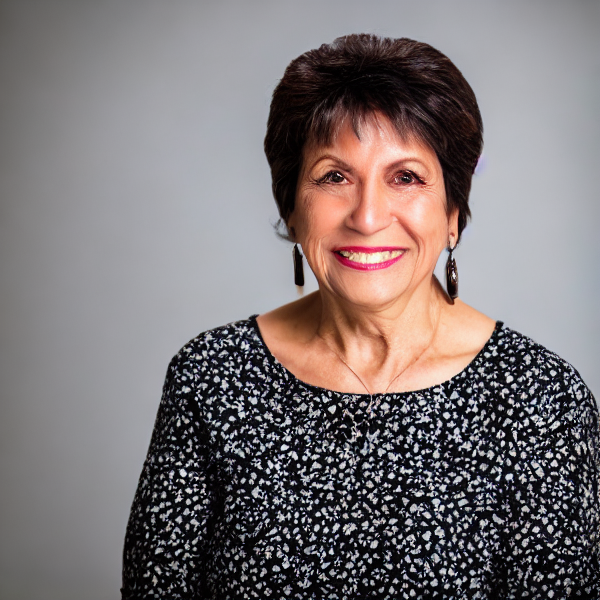} &
        \includegraphics[width=0.11\textwidth]{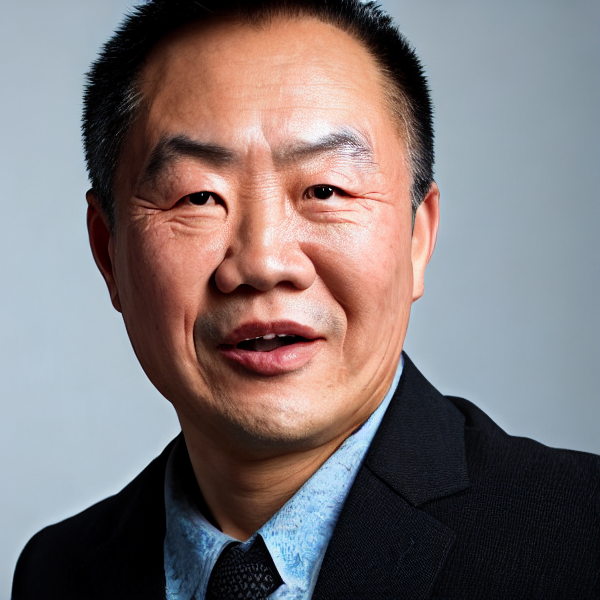} &
        \includegraphics[width=0.11\textwidth]{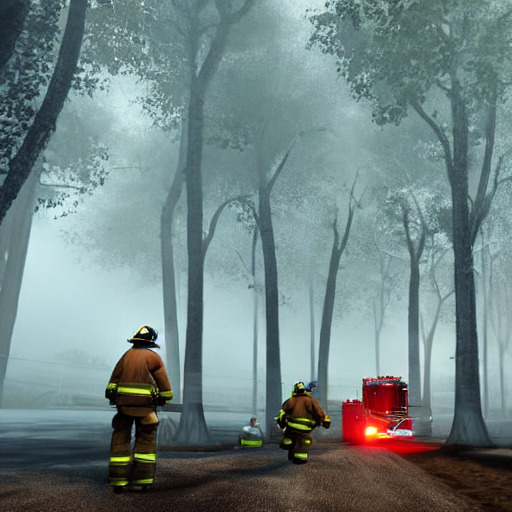} &
        \includegraphics[width=0.11\textwidth]{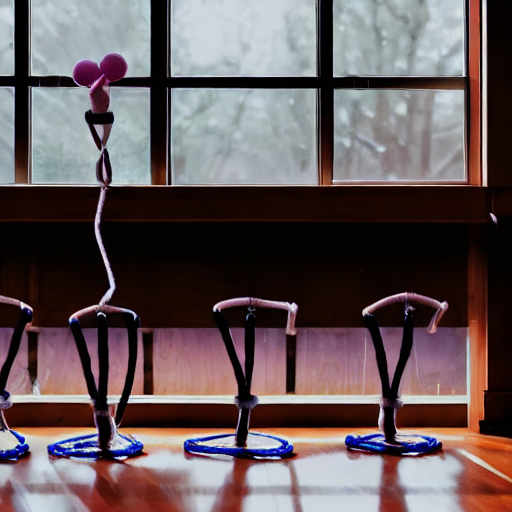} &
        \includegraphics[width=0.11\textwidth]{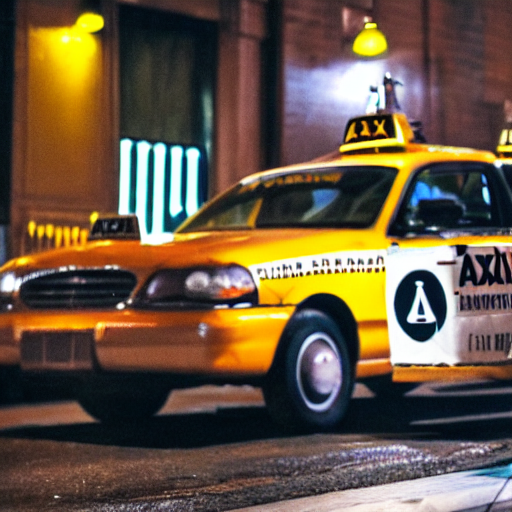}
        \\
        \raisebox{0.9cm}{Stable Diffusion v2} & 
        \includegraphics[width=0.11\textwidth]{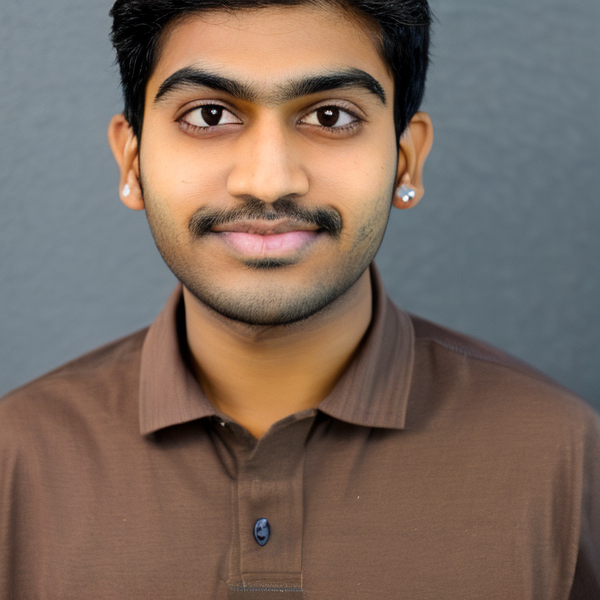} &
        \includegraphics[width=0.11\textwidth]{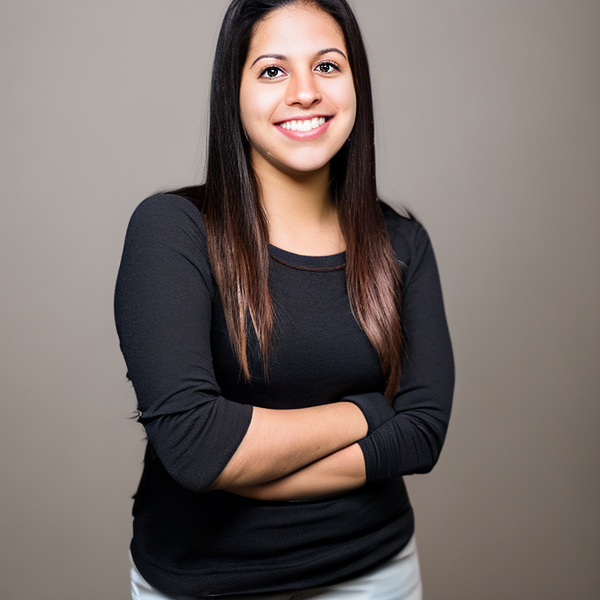} &
        \includegraphics[width=0.11\textwidth]{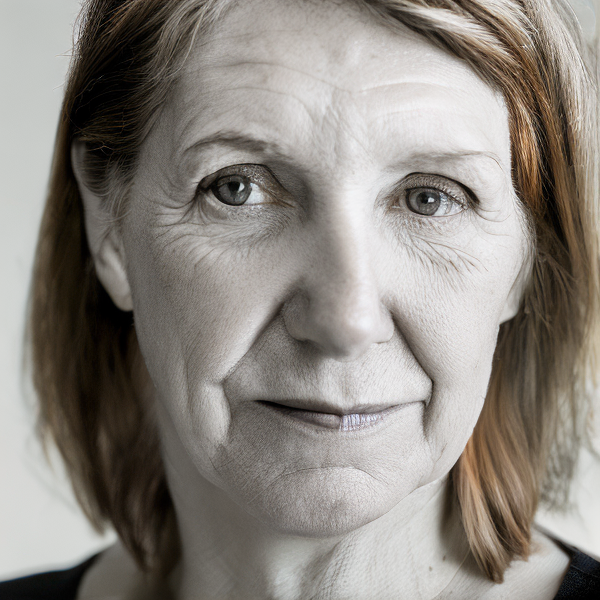} &
        \includegraphics[width=0.11\textwidth]{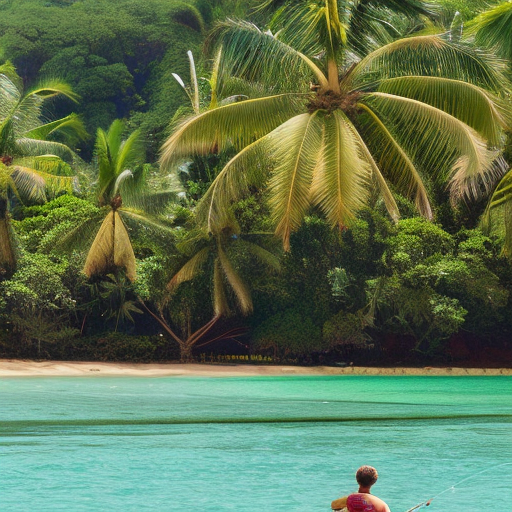} &
        \includegraphics[width=0.11\textwidth]{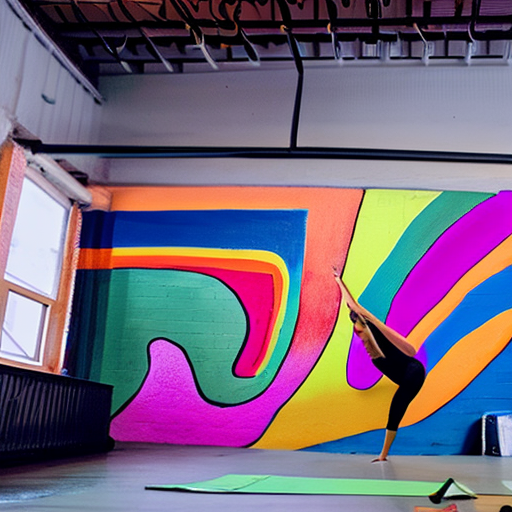} &
        \includegraphics[width=0.11\textwidth]{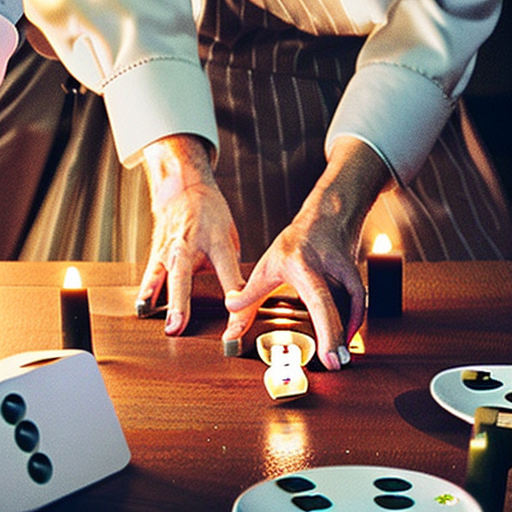}
        \\
        \raisebox{0.9cm}{Stable Diffusion xl} & 
        \includegraphics[width=0.11\textwidth]{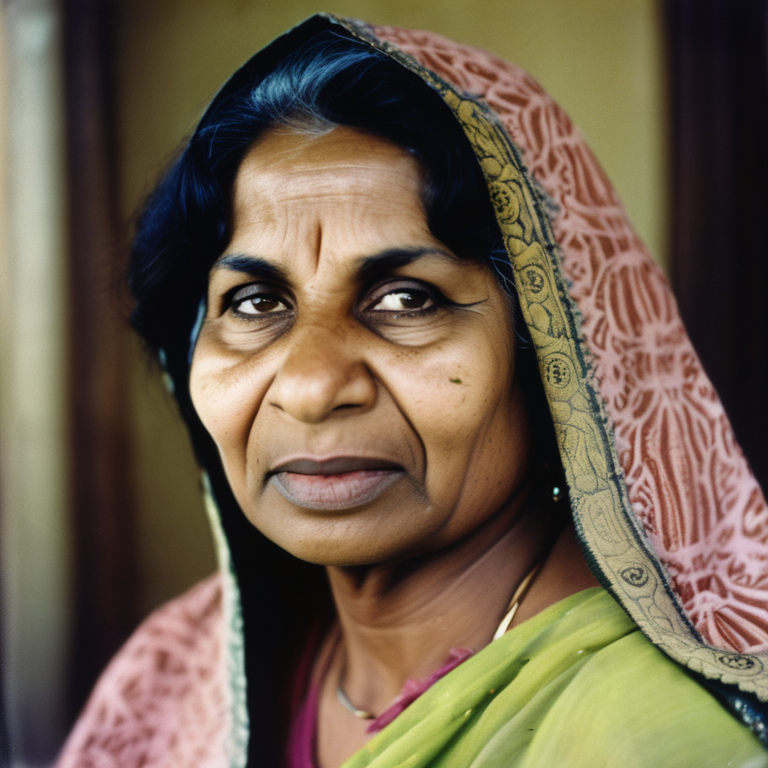} &
        \includegraphics[width=0.11\textwidth]{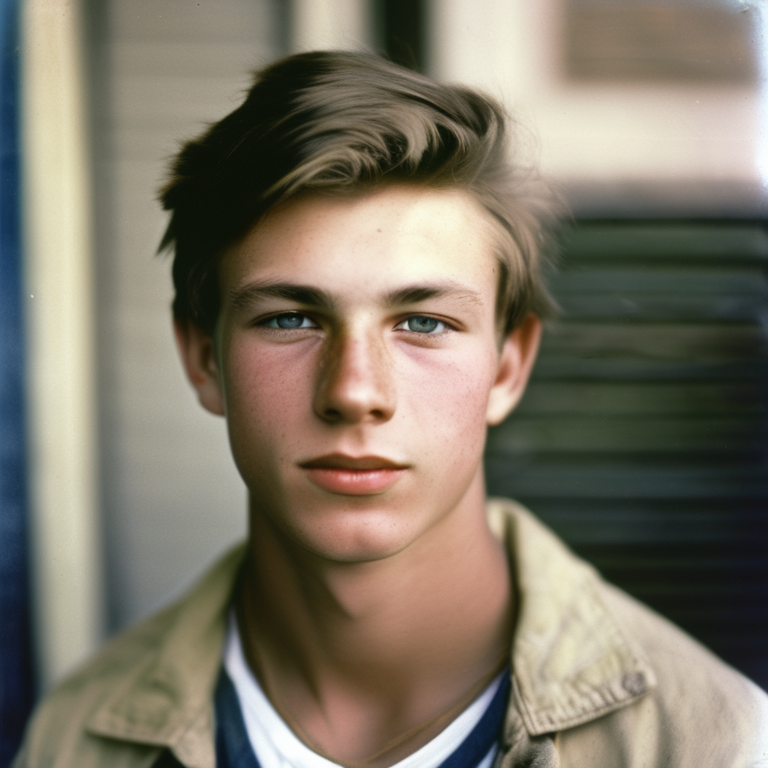} &
        \includegraphics[width=0.11\textwidth]{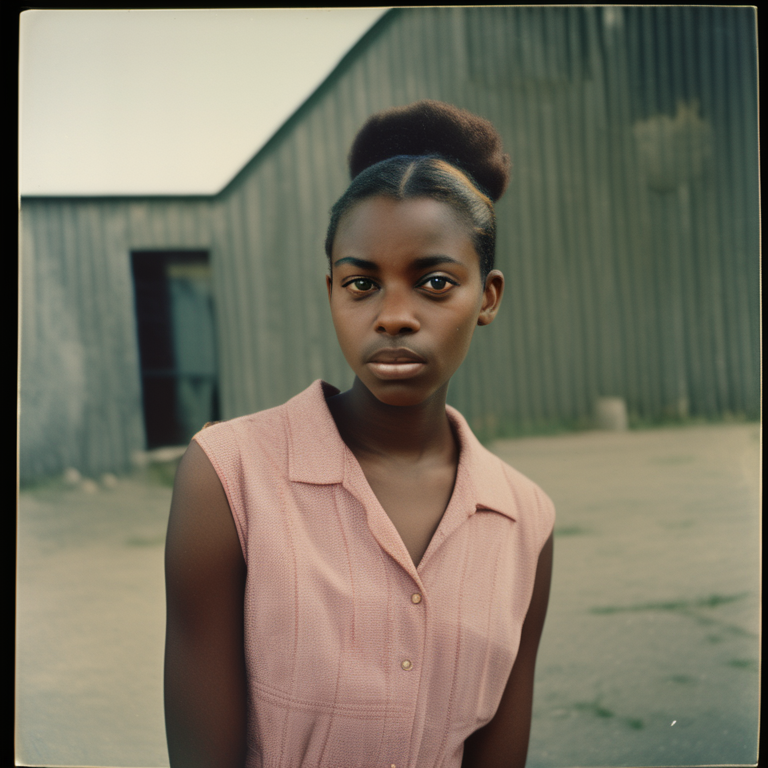} &
        \includegraphics[width=0.11\textwidth]{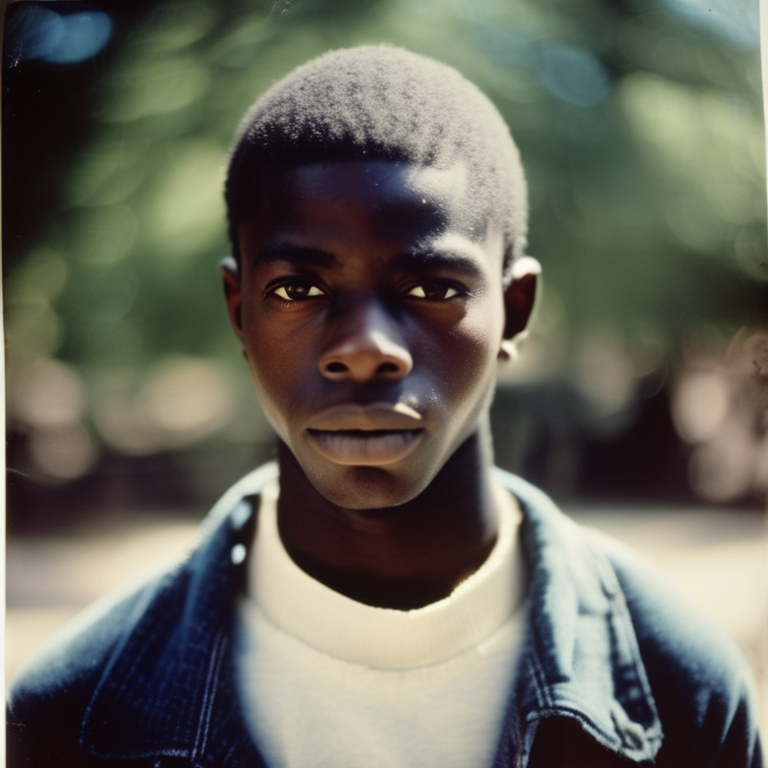} &
        \includegraphics[width=0.11\textwidth]{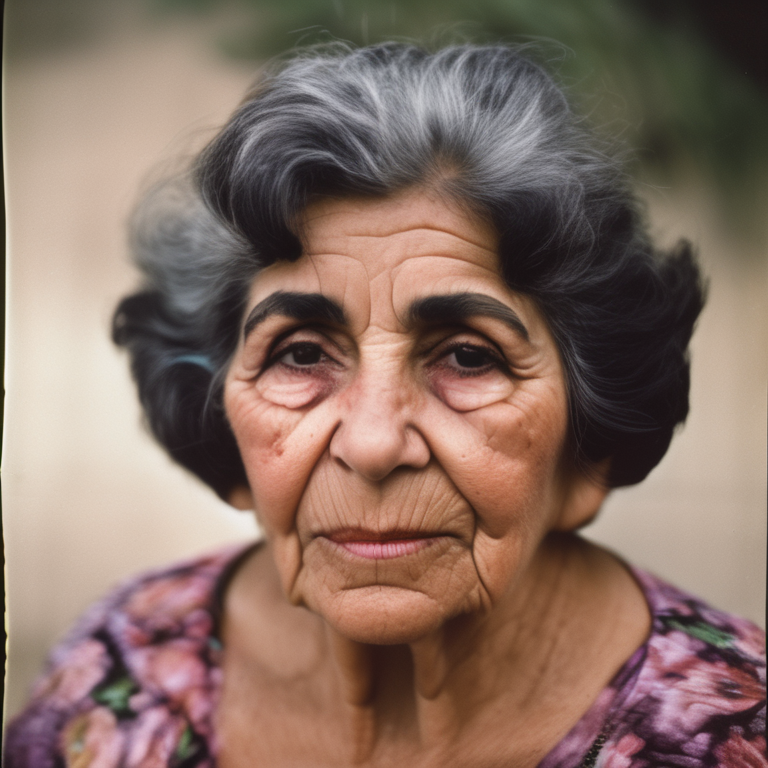} &
        \includegraphics[width=0.11\textwidth]{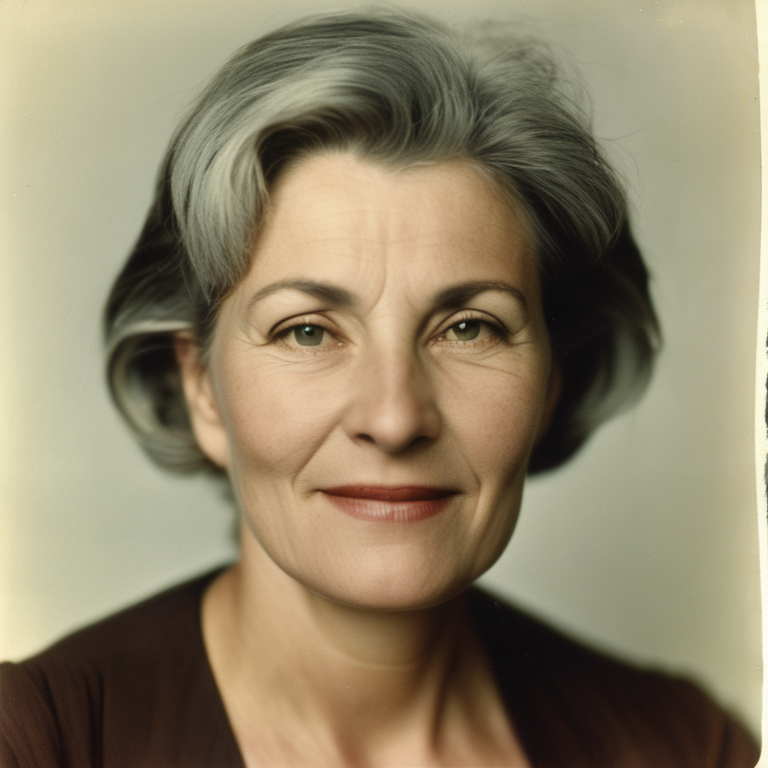}
    \end{tabular}
    \end{center}
    \vspace{-0.5cm}
    \caption{Representative examples of AI-generated images used in our training and evaluation (see also Table~\ref{tab:datasets}). Some synthesis engines were used to generate faces only and others were used to synthesize both faces and non-faces. In order to respect user privacy, we do not show examples of real photos.}
    \label{fig:datasets}
\end{figure*}

Our training and evaluation leverage $18$ data sets consisting of $120{\small ,}000$ real LinkedIn profile photos and $105{\small ,}900$ AI-generated faces spanning five different GAN and five different diffusion synthesis engines. The AI-generated images consist of two main categories, those with a face and those without. Real and synthesized color (RGB) images are resized from their original resolution to $512 \times 512$ pixels. Shown in Table~\ref{tab:datasets} is an accounting of these images, and shown in Figure~\ref{fig:datasets} are representative examples from each of the AI-generated categories as described next.

\begin{table}[t]
    \centering
    \resizebox{0.475\textwidth}{!}{
    \begin{tabular}{r|ccr}
    {\bf type} & {\bf model} & {\bf category} &  {\bf number} \\
    \hline
    real         & -                   & faces    & $120{\small ,}000$ \\
    \hline
    AI-GAN       & generated.photos    & faces    & $10{\small ,}000$ \\
    AI-GAN       & StyleGAN 1          & faces    & $10{\small ,}000$ \\
    AI-GAN       & StyleGAN 2          & faces    & $10{\small ,}000$ \\
    AI-GAN       & StyleGAN 3          & faces    & $10{\small ,}000$ \\
    AI-GAN       & EG3D                & faces    & $10{\small ,}000$ \\
    AI-GAN       & StyleGAN 1          & cars     & $5{\small ,}000$ \\
    AI-GAN       & StyleGAN 1          & bedrooms & $5{\small ,}000$ \\
    AI-GAN       & StyleGAN 1          & cats     & $5{\small ,}000$ \\
    \hline-
    AI-diffusion & DALL-E 2            & faces    & $9{\small ,}000$ \\
    AI-diffusion & Midjourney          & faces    & $9{\small ,}000$ \\
    AI-diffusion & Stable Diffusion 1  & faces    & $9{\small ,}000$ \\
    AI-diffusion & Stable Diffusion 2  & faces    & $9{\small ,}000$ \\
    AI-diffusion & Stable Diffusion xl & faces    & $900$ \\
    AI-diffusion & DALL-E 2            & random   & $1{\small ,}000$ \\
    AI-diffusion & Midjourney          & random   & $1{\small ,}000$ \\
    AI-diffusion & Stable Diffusion 1  & random   & $1{\small ,}000$ \\
    AI-diffusion & Stable Diffusion 2  & random   & $1{\small ,}000$ \\    
    \hline
    \end{tabular}
    }
    \vspace{-0.1cm}
    \caption{A breakdown of the number of real and AI-generated images used in our training and evaluation (see also Figure~\ref{fig:datasets}).}
    \label{tab:datasets}
\end{table}

\subsection{Real Faces}
\label{subsec:real-faces}

The $120{\small ,}000$ real photos were sampled from LinkedIn members with publicly-accessible profile photos uploaded between January 1, 2019 and December 1, 2022. These accounts showed activity on the platform on at least $30$ days (e.g.,~signed in, posted, messaged, searched) without triggering any fake-account detectors. Given the age and activity on the accounts, we can be confident that these photos are real. These images were of widely varying resolution and quality. Although most of these images are standard profile photos consisting of a single person, some do not contain a face. In contrast, all of the AI-generated images (described next) consist of a face. We will revisit this difference between real and fake images in Section~\ref{sec:results}.

\subsection{GAN Faces}

We generated a total of $10{\small,}000$ images from each StyleGAN version (1, 2, 3, EG3D)~\cite{karras2019style,karras2020analyzing,karras2021alias,chan2022efficient}. For versions 1, 2, and 3, color images were synthesized at a resolution of $1024 \times 1024$ pixels and with $\psi=0.5$.\footnote{The StyleGAN parameter $\psi$ (typically in the range $[0,1]$) controls the truncation of the seed values in the latent space representation used to generate an image. Smaller values of $\psi$ provide better image quality but reduce facial variety. A mid-range value of $\psi=0.5$ produces relatively artifact-free faces, while allowing for variation in the gender, age, and ethnicity in the synthesized face.} For EG3D (Efficient Geometry-aware 3D Generative Adversarial Networks), the so-called 3D version of StyleGAN, we synthesized $10{\small,}000$ images at a resolution of $512 \times 512$, with $\psi=0.5$, and with random head poses.

A total of $10{\small,}000$ images at a resolution of $1024 \times 1024$ pixels were downloaded from generated.photos\footnote{\url{https://generated.photos/faces}}. These GAN-synthesized images generally produce more professional looking head shots because the network is trained on a dataset of high-quality images recorded in a photographic studio.

\subsection{GAN Non-Faces}

A total of $5{\small,}000$ StyleGAN 1 images were downloaded\footnote{\url{https://github.com/NVlabs/stylegan)}} for each of three non-face categories: bedrooms, cars, and cats (the repositories for other StyleGAN versions do not provide images for categories other than faces). These images ranged in size from $512 \times 384$ (cars) to $256 \times 256$ (bedrooms and cats).

\subsection{Diffusion Faces}

We generated $9{\small,}000$ images from each Stable Diffusion~\cite{rombach2022high} version (1, 2)\footnote{\url{https://github.com/Stability-AI/StableDiffusion}}. Unlike the GAN faces described above, text-to-image diffusion synthesis affords more control over the appearance of the faces. To ensure diversity, $300$ faces for each of $30$ demographics with the prompts ``a photo of a \{young, middle-aged, older\} \{black, east-asian, hispanic, south-asian, white\} \{woman, man\}.'' These images were synthesized at a resolution of $512 \times 512$. This dataset was curated to remove obvious synthesis failures in which, for example, the face was not visible. 

An additional $900$ images were synthesized from the most recent version of Stable Diffusion (xl). Using the same demographic categories as before, $30$ images were generated for each of $30$ categories, each at a resolution of $768 \times 768$.

We generated $9{\small,}000$ images from DALL-E 2\footnote{\url{https://openai.com/dall-e-2}}, consisting of $300$ images for each of $30$ demographic groups. These images were synthesized at a resolution of $512 \times 512$ pixels.

A total of $1{\small ,}000$ Midjourney\footnote{\url{https://www.midjourney.com}} images were downloaded at a resolution of $512 \times 512$. These images were manually curated to consist of only a single face.

\subsection{Diffusion Non-Faces}

We synthesized $1{\small ,}000$ non-face images from each of two versions of Stable Diffusion (1, 2). These images were generated using random captions (generated by ChatGPT) and were manually reviewed to remove any images containing a person or face. These images were synthesized at a resolution of $600 \times 600$ pixels. A similar set of $1{\small ,}000$ DALL-E 2 and $1{\small ,}000$ Midjourney images were synthesized at a resolution of $512 \times 512$.

\subsection{Training and Evaluation Data}
\label{subsec:training-and-evaluation}

The above enumerated sets of images are split into training and evaluation as follows. Our model (described in Section~\ref{sec:model}) is trained on a random subset of $30{\small ,}000$ real faces and $30{\small ,}000$ AI-generated faces. The AI-generated faces are comprised of a random subset of $5{\small ,}250$ StyleGAN 1, $5{\small ,}250$ StyleGAN 2, $4{\small ,}500$ StyleGAN 3, $3{\small ,}750$ Stable Diffusion 1, $3{\small ,}750$ Stable Diffusion 2, and $7{\small ,}500$ DALL-E 2 images.

We evaluate our model against the following:
\begin{itemize}
    \setlength\itemsep{0.1em}
    \item A set of $5{\small ,}000$ face images from the same synthesis engines used in training (StyleGAN 1, StyleGAN 2, StyleGAN 3, Stable Diffusion 1, Stable Diffusion 2, and DALL-E 2). 
    \item A set of $5{\small ,}000$ face images from synthesis engines not used in training (Generated.photos, EG3D, Stable Diffusion xl, and Midjourney). 
    \item A set of $3{\small ,}750$ non-face images from each of five synthesis engines (StyleGAN 1, DALL-E 2, Stable Diffusion 1, Stable Diffusion 2, and Midjourney).
    \item A set of $13{\small ,}750$ real faces.
\end{itemize}
%
%

\section{Model}
\label{sec:model}

We train a model to distinguish real from AI-generated faces. The underlying model is the EfficientNet-B1\footnote{We are describing an older version of the EfficientNet model which we have previously operationalized on LinkedIn that has since been replaced with a new model. We recognize that this model is not the most recent, but we are only now able to report these results since the model is no longer in use.} convolutional neural network~\cite{tan2020efficientnet}. We found that this architecture provides better performance as compared to other state-of-the-art architectures (Swin-T~\cite{liu2021Swin}, Resnet50~\cite{he2015deep}, XceptionNet~\cite{chollet2017xception}). The EfficientNet-B1 network has $7.8$ million internal parameters that were pre-trained on the ImageNet-1K image dataset~\cite{tan2020efficientnet}.

Our pipeline consists of three stages: (1) an image pre-processing stage; (2) an image embedding stage; and (3) a scoring stage. The model takes as input a color image and generates a numerical score in the range $[0,1]$. Scores near $0$ indicate that the image is likely real, and scores near $1$ indicate that the image is likely AI-generated.

The image pre-processing step resizes the input image to a resolution of $512 \times 512$ pixels. This resized color image is then passed to an EfficientNet-B1 transfer learning layer. In the scoring stage, the output of the transfer learning layer is fed to two fully connected layers, each of size $2{\small ,}048$, with a ReLU activation function, a dropout layer with a $0.8$ dropout probability, and a final scoring layer with a sigmoidal activation. Only the scoring layers -- with $6.8$ million trainable parameters -- are tuned. The trainable weights are optimized using the AdaGrad algorithm with a mini-batch of size $32$, a learning rate of $0.0001$, and trained for up to $10{\small , }000$ steps. A cluster with $60$ NVIDIA A100 GPUs was used for model training.

\section{Results}
\label{sec:results}

\begin{table}[t]
    \begin{center}
    \resizebox{0.475\textwidth}{!}{
    \begin{tabular}{r|c *{1}{d{8.1}} *{1}{d{6.1}}}
        \mc{\bf condition} & \mc{\bf image} & \mc{\bf TPR} &\mc{\bf F1} \\
        \hline
        training                   & face     & $100.0\%$ & $0.998$ \\
        evaluation (in-engine)     & face     & $98.0\%$  & $0.987$ \\
        evaluation (out-of-engine) & face     & $84.5\%$  & $0.914$ \\
        evaluation (in/out-engine) & non-face & $0.0\%$   & $0.000$ \\
        \hline
    \end{tabular}
    }
    \end{center}
    \vspace{-0.5cm}
    \caption{Baseline training and evaluation true positive (correctly classifying an AI-generated image, averaged across all synthesis engines (TPR)). In each condition, the false positive rate is $0.5\%$ (incorrectly classifying a real face (FPR)). Also reported is the F1 score defined as $2\textrm{TP}/(2\textrm{TP}+\textrm{FP}+\textrm{FN})$. TP, FP, and FN represent the number of true positives, false positives, and false negatives, respectively. In-engine/out-of-engine indicates that the images were created with the same/different synthesis engines as those used in training.}
    \label{tab:results}
\end{table}

Our baseline training and evaluation performance is shown in Table~\ref{tab:results}. The evaluation is broken down based on whether the evaluation images contain a face or not (training images contained only faces) and whether the images were generated with the same (in-engine) or different (out-of-engine) synthesis engines as those used in training (see Section~\ref{subsec:training-and-evaluation}). In order to provide a direct comparison of the true positive rate\footnote{True positive rate (TPR) is the fraction of AI-generated photos that are correctly classified.} (TPR) for the training and evaluation, we adjust the final classification threshold to yield a false positive rate\footnote{False positive rate (FPR) is the fraction of real photos that are incorrectly classified.} (FPR) of $0.5\%$.

\begin{figure}[t]
    \begin{tabular}{c@{\hspace{0.1cm}}c}
    \raisebox{3.6cm}{(a)} & 
    \includegraphics[width=0.42\textwidth]{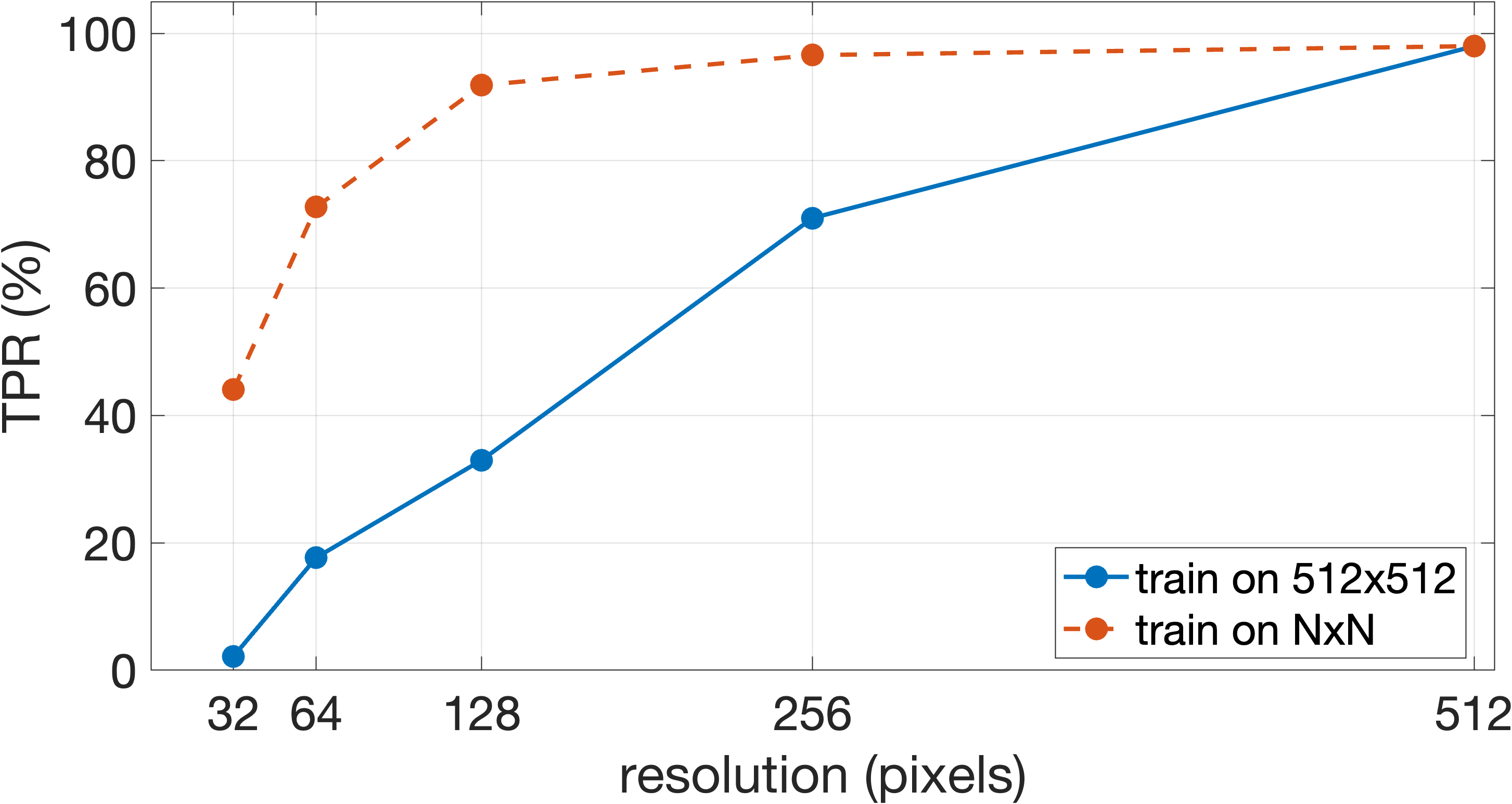} \\
    \\
    \raisebox{3.6cm}{(b)} & 
    \includegraphics[width=0.42\textwidth]{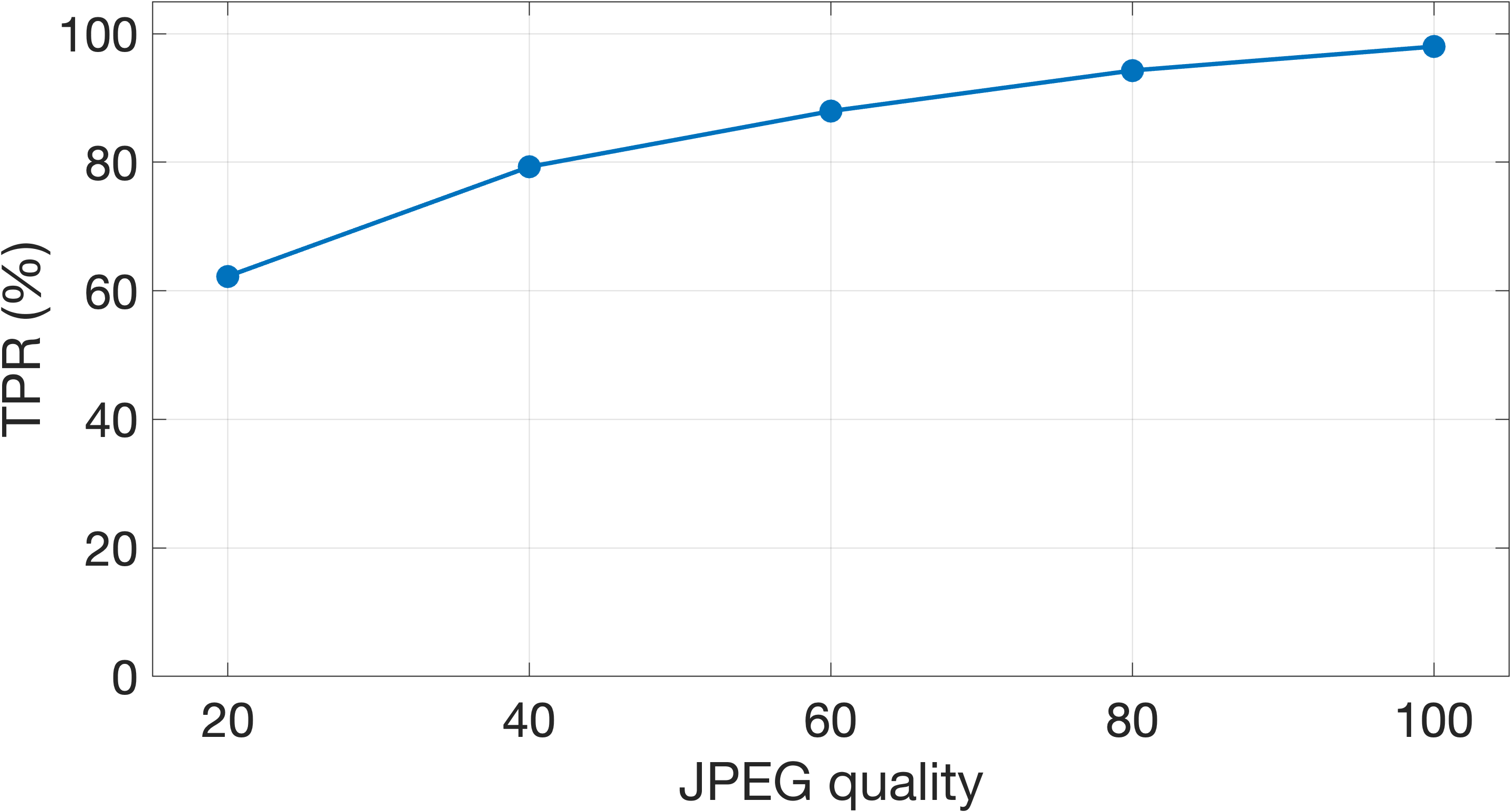}
    \end{tabular}
    \vspace{-0.25cm}
\caption{True positive rate (TPR) for correctly classifying an AI-generated face (with a fixed FPR of $0.5\%$) as a function of: (a) resolution where the model is trained on $512 \times 512$ images and evaluated against different resolution (solid blue) and trained and evaluated on a single resolution $N \times N$ (dashed red); and (b) JPEG quality where the model is trained on uncompressed images and a range of JPEG compressed images and evaluated on JPEG qualities between $20$ (lowest) to $100$ (highest).}
\label{fig:results-res-comp}
\end{figure}

With a fixed FPR of $0.5\%$, AI-generated faces are correctly classified in training and evaluation at a rate of $98\%$. Across different synthesis engines (StyleGAN 1,2,3, Stable Diffusion 1,2, and DALL-E 2) used for training, TPR was varied somewhat from a low of $93.3\%$ for Stable Diffusion 1 to a high of $99.5\%$ for StyleGAN 2, and $98.9\%$ for StyleGAN1, $99.9\%$ for StyleGAN3, $94.9\%$ for Stable Diffusion 2, and $99.2\%$ for DALL-E 2.

For faces generated by synthesis engines not used in training (out-of-engine), TPR drops to $84.5\%$ at the same FPR, showing good but not perfect out-of-domain generalization. Across the different synthesis engines not used in training, TPR varied widely with a low of $19.4\%$ for Midjourney to a high of $99.5\%$ for EG3D, and $95.4\%$ for generated.photos. Our classifier generalizes well in some cases, and fails in others. This limitation, however, can likely be mitigated by incorporating these out-of-engine images into the initial training.

In a particularly striking result, non-faces -- generated by the same synthesis engines as used in training -- are all incorrectly classified. This is most likely because some of our real images contain non-faces (see Section~\ref{subsec:real-faces}) while all of the AI-generated images contain faces. Since we are only interested in detecting fake faces used to create an account, we don't see this as a major limitation. This result also suggests that our classifier has latched onto a specific property of an AI-generated face and not some low-level artifact from the underlying synthesis (e.g.,~a noise fingerprint~\cite{corvi2023detection}). In Section~\ref{subsec:explainability}, we provide additional evidence to support this hypothesis.

The above baseline results are based on training and evaluating images at a resolution of $512 \times 512$ pixels. Shown in Figure~\ref{fig:results-res-comp}(a) (solid blue) is the TPR when the training images are down scaled to a lower resolution ($256$, $128$, $64$, and $32$) and then up scaled back up to $512$ for classification. With the same FPR of $0.5\%$, TPR for classifying an AI-generated face drops fairly quickly from a baseline of $98.0\%$. 

The true positive rate, however, improves significantly when the model is trained on images at a resolution of $N \times N$ ($N = 32, 64, 128,$ or $256$) and then evaluated against the same TPR seen in training, Figure~\ref{fig:results-res-comp}(a) (dashed red). As before, the false positive rate is fixed at $0.5\%$. Here we see that TPR at a resolution of $128 \times 128$ remains relatively high ($91.9\%$) and only degrades at the lowest resolution of $32 \times 32$ ($44.1\%$). The ability to detect AI-generated faces at even relatively low resolutions suggests that our model has not latched onto a low-level artifact that would not survive this level of down-sampling.

Shown in Figure~\ref{fig:results-res-comp}(b) is the TPR of the classifier, trained on uncompressed PNG and JPEG images of varying quality, evaluated against images across a range of JPEG qualities (ranging from the highest quality of $100$ to the lowest quality of $20$). Here we see that TPR for identifying an AI-generated face (FPR is $0.5\%$) degrades slowly with a TPR of $94.3\%$ at quality $80$ and a TPR of $88.0\%$ at a quality of $60$. Again, the ability to detect AI-generated faces in the presence of JPEG compression artifacts suggests that our model has not latched onto a low-level artifact.

\subsection{Explainability}
\label{subsec:explainability}

As shown in Section~\ref{sec:results}, our classifier is highly capable of distinguishing AI faces generated from a range of different synthesis engines. This classifier, however, is limited to only faces, Table~\ref{tab:results}. That is, when presented with non-face images from the same synthesis engines as used in training, the classifier completely fails to classify them as AI-generated. 

We posit that our classifier may have learned a semantic-level artifact. This claim is partly supported by the fact that our classifier remains highly accurate even at resolutions as low as $128 \times 128$ pixels, Figure~\ref{fig:results-res-comp}(a), and remains reasonably accurate even in the face of fairly aggressive JPEG compression, Figure~\ref{fig:results-res-comp}(b). Here we provide further evidence to support this claim that we have learned a structural- or semantic-level artifact. 

\begin{figure}[t]
    \includegraphics[width=0.475\textwidth]{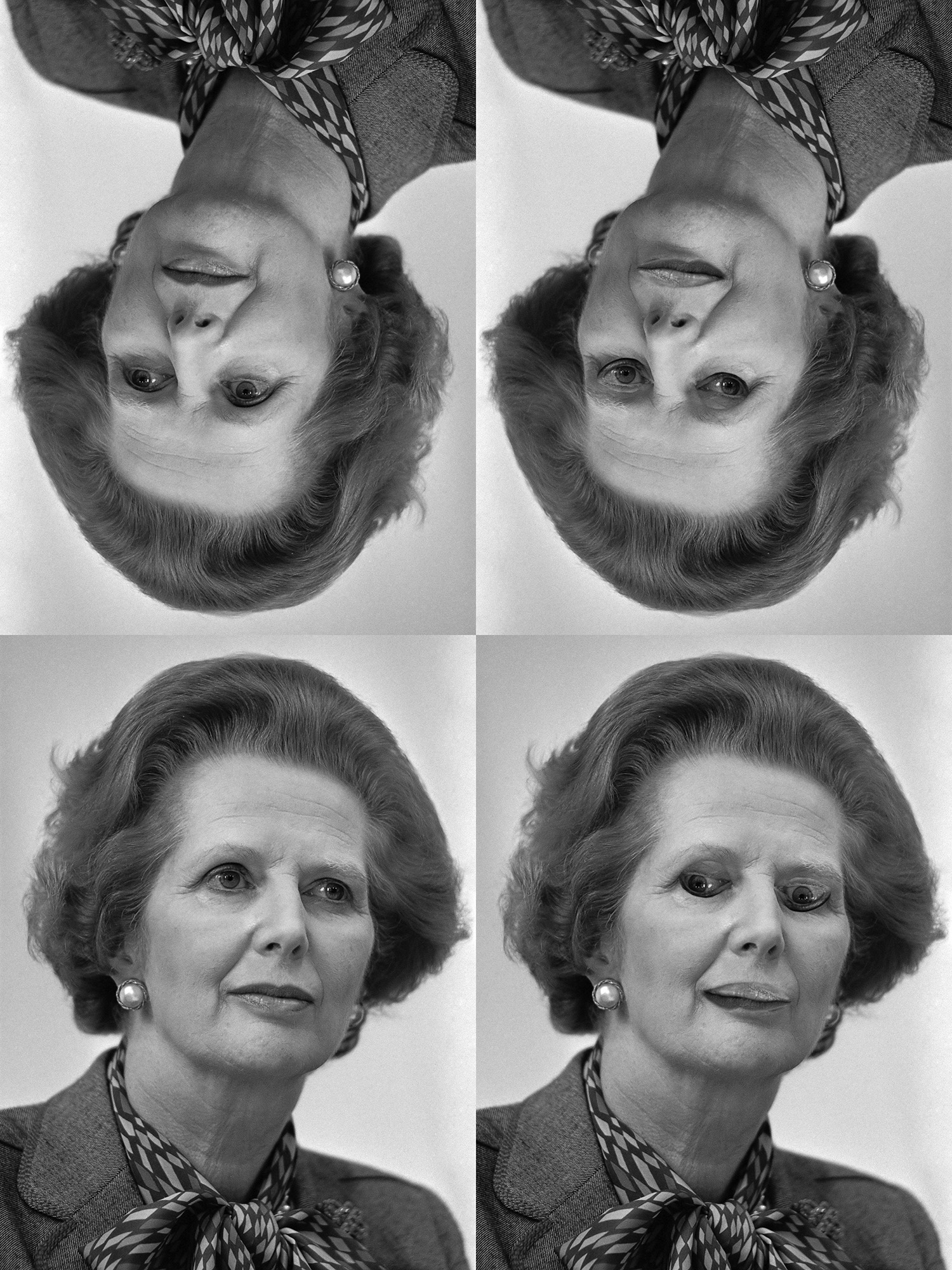}
    \vspace{-0.5cm}
    \caption{The Margaret Thatcher illusion~\cite{thompson1980margaret}: the faces in the top row are inverted versions of those on the bottom row. The eye and mouth inversion in the bottom right is evident when the face is upright, but not when it is inverted. (Credit: Rob Bogaerts/Anefo  \url{https://commons.wikimedia.org/w/index.php?curid=79649613)})}
    \label{fig:thatcher}
\end{figure}

It is well established that while general-purpose object recognition in the human visual system is highly robust to object orientation, pose, and perspective distortion, face recognition and processing are less robust to even a simple inversion~\cite{sinha2006face}. This effect is delightfully illustrated in the classic Margaret Thatcher illusion~\cite{thompson1980margaret}. The faces in the top row of Figure~\ref{fig:thatcher} are inverted versions of those in the bottom row. In the version on the right, the eyes and mouth are inverted relative to the face. This grotesque feature cocktail is obvious in the upright face but not in the inverted face.

We wondered if our classifier would struggle to classify vertically inverted faces. The same $10{\small ,}000$ validation images (Section~\ref{subsec:training-and-evaluation}) were inverted and re-classified. With the same fixed FPR of $0.5\%$, TPR dropped by $20$ percentage points from $98.0\%$ to $77.7\%$. 

By comparison, flipping the validation images about just the vertical axis (i.e.,~left-right flip) yields no change in the TPR of $98.0\%$ with the same $0.5\%$ FPR. This pair of results, combined with the robustness to resolution and compression quality, suggest that our model has not latched onto a low-level artifact, and may have instead discovered a structural or semantic property that distinguishes AI-generated faces from real faces.

We further explore the nature of our classifier using the method of integrated gradients~\cite{sundararajan2017axiomatic}. This method attributes the predictions made by a deep network to its input features. Because this method can be applied without any changes to the trained model, it allows us to compute the relevance of each input image pixel with respect to the model's decision.

\begin{figure}[t]
    \begin{center}
        \begin{tabular}{r@{\hspace{0.15cm}}c@{\hspace{0.15cm}}c}
            & image & gradients \\
            \raisebox{1.6cm}{(a)} & 
            \includegraphics[width=0.2\textwidth]{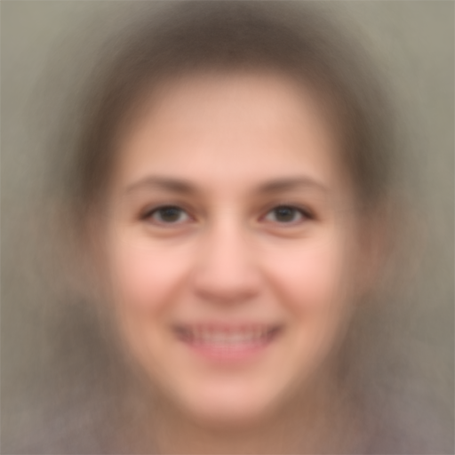} & 
            \includegraphics[width=0.2\textwidth]{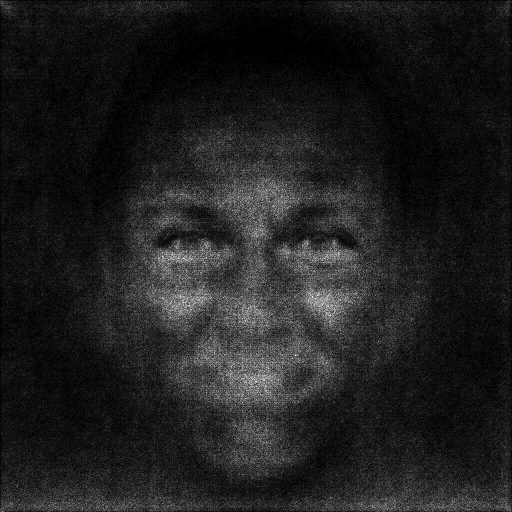} \\
            \raisebox{1.6cm}{(b)} &  
             \includegraphics[width=0.2\textwidth]{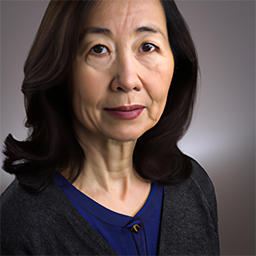} & 
             \includegraphics[width=0.2\textwidth]{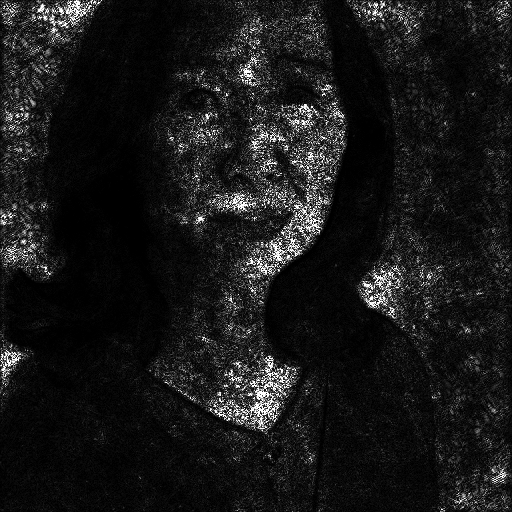} \\
            \raisebox{1.6cm}{(c)} & 
             \includegraphics[width=0.2\textwidth]{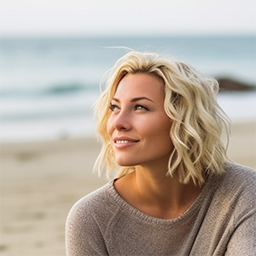} &
             \includegraphics[width=0.2\textwidth]{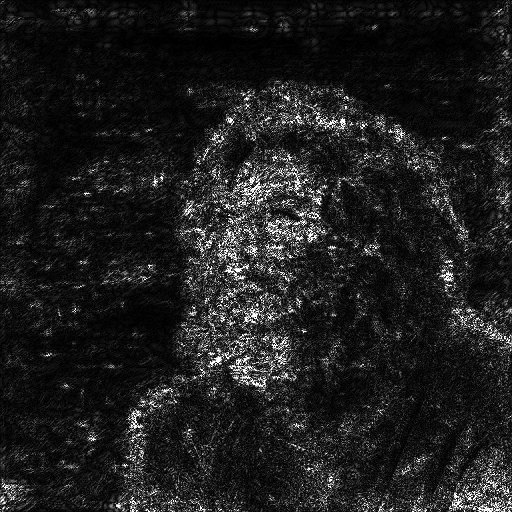}
             \\
            \raisebox{1.6cm}{(d)} & 
            \includegraphics[width=0.2\textwidth]{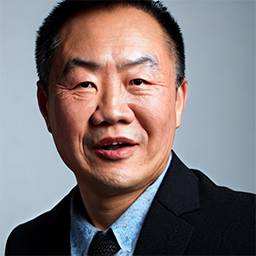} & 
             \includegraphics[width=0.2\textwidth]{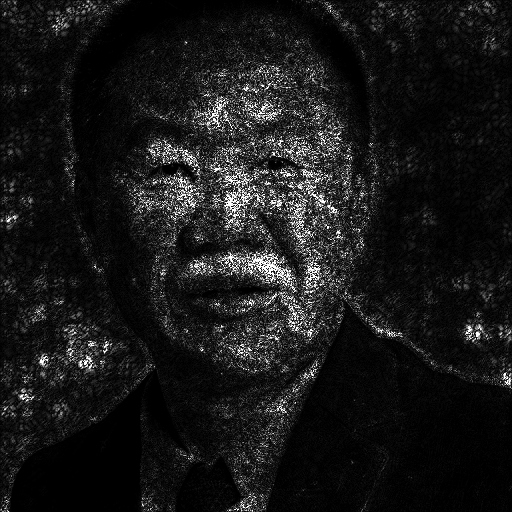}
            \\
            \raisebox{1.6cm}{(e)} & 
            \includegraphics[width=0.2\textwidth]{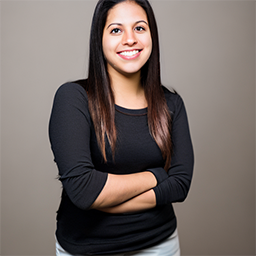} & 
             \includegraphics[width=0.2\textwidth]{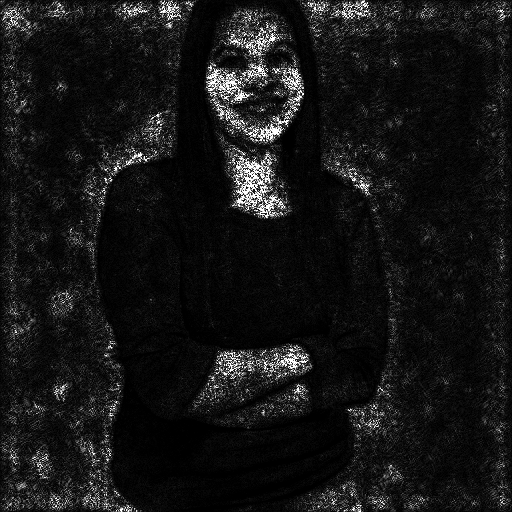}
            \\
            
        \end{tabular}
    \end{center}
    \vspace{-0.5cm}
    \caption{Examples of AI-generated faces and their normalized integrated gradients, revealing that our model is primarily focused on facial regions: (a) an average of $100$ StyleGAN 2 faces, (b) DALL-E 2, (c) Midjourney, (d,e) Stable Diffusion 1,2.}
    \label{fig:xAI}
\end{figure}

Shown in Figure~\ref{fig:xAI}(a) is the unsigned magnitude of the normalized (into the range $[0,1]$) integrated gradients averaged over $100$ StyleGAN 2 images (because the StyleGAN-generated faces are all aligned, the averaged gradient is consistent with facial features across all images). Shown in Figure~\ref{fig:xAI}(b)-(e) are representative images and their normalized integrated gradients for an image generated by DALL-2, Midjourney, Stable Diffusion 1, and Stable Diffusion 2. In all cases, we see that the most relevant pixels, corresponding to larger gradients, are primarily focused around the facial region and other areas of skin.

\subsection{Comparison}
\label{sec:comparison}

Because it focused specifically on detecting GAN-generated faces, the work of~\cite{mundra2023exposing} is most directly related to ours. In this work, the authors show that a low-dimensional linear model captures the common facial alignment of StyleGAN-generated faces. Evaluated against $3{\small, }000$ StyleGAN faces, their model correctly classifies $99.5\%$ of the GAN faces with $1\%$ of real faces incorrectly classified as AI. By comparison, we achieve a similar TPR, but with a lower $0.5\%$ FPR. 

Unlike our approach, however, which generalizes to other GAN faces like generated.photos, TPR for this earlier work drops to $86.0\%$ (with the same $1\%$ FPR). Furthermore, this earlier work fails to detect diffusion-based faces because these faces simply do not contain the same alignment artifact as StyleGAN faces. By comparison, our technique generalizes across GAN- and diffusion-generated faces and to synthesis engines not seen in training.

We also evaluated a recent state-of-the-art model that exploits the presence of Fourier artifacts in AI-generated images~\cite{corvi2023detection}. On our evaluation dataset of real and in-engine AI-generated faces this model correctly classifies only $23.8\%$ of the AI-generated faces at the same FPR of $0.5\%$. This TPR is considerably lower than our model's TPR of $98.0\%$ and also lower than ~$90\%$ TPR reported in~\cite{corvi2023detection}. We hypothesize that this discrepancy is due to the more diverse and challenging in-the-wild real images of our dataset.


\section{Discussion}
\label{sec:discussion}

For many image classification problems, large neural models -- with appropriately representative data -- are attractive for their ability to learn discriminating features. These models, however, can be vulnerable to adversarial attacks~\cite{carlini2017towards}. It remains to be seen if our model is as vulnerable as previous models in which imperceptible amounts of adversarial noise confound the model~\cite{carlini2020evading}. In particular, it remains to be seen if the apparent structural or semantic artifacts we seem to have learned will yield more robustness to intentional adversarial attacks.

In terms of less sophisticated attacks, including laundering operations like transcoding and image resizing, we have shown that our model is resilient across a broad range of laundering operations.

The creation and detection of AI-generated content is inherently adversarial with a somewhat predictable back and forth between creator and detector. While it may seem that detection is futile, it is not. By continually building detectors, we force creators to continue to invest time and cost to create convincing fakes. And while the sufficiently sophisticated creator will likely be able to bypass most defenses, the average creator will not. 

When operating on large online platforms like ours, this mitigation -- but not elimination -- strategy is valuable to creating safer online spaces. In addition, any successful defense will employ not one, but many different approaches that exploit various artifacts. Bypassing all such defenses will pose significant challenges to the adversary. By learning what appears to be a robust artifact that is resilient across resolution, quality, and a range of synthesis engines, the approach described here adds a powerful new tool to a defensive toolkit.


\section*{Acknowledgements}
\label{sec:acknowledgements}

This work is the product of a collaboration between Professor Hany Farid and the Trust Data team at LinkedIn\footnote{The model described in this work is not used to take action on any LinkedIn members.}. We thank Maty{\'a}{\v{s}} Boh{\'a}{\v{c}}ek for his help in creating the AI-generated faces. We thank the LinkedIn Scholars\footnote{\url{https://careers.linkedin.com/scholars}} program for enabling this collaboration. We also thank Ya Xu, Daniel Olmedilla, Kim Capps-Tanaka, Jenelle Bray, Shaunak Chatterjee, Vidit Jain, Ting Chen, Vipin Gupta, Dinesh Palanivelu, Milinda Lakkam, and Natesh Pillai for their support of this work. We are grateful to David Luebke, Margaret Albrecht, Edwin Nieda, Koki Nagano, George Chellapa, Burak Yoldemir, and Ankit Patel at NVIDIA for facilitating our work by making the StyleGAN generation software, trained models and synthesized images publicly available, and for their valuable suggestions.


{\small
\bibliographystyle{ieee_fullname}
\bibliography{main}
}

\end{document}